\documentclass[runningheads]{llncs}

\usepackage{eccv}

\usepackage{eccvabbrv}

\usepackage{amsmath,amsfonts,bm}

\def\eqref#1{equation~\ref{#1}}

\def\1{\bm{1}}

\def\vtheta{{\bm{\theta}}}

\def\vx{{\bm{x}}}

\def\vz{{\bm{z}}}

\DeclareMathAlphabet{\mathsfit}{\encodingdefault}{\sfdefault}{m}{sl}
\SetMathAlphabet{\mathsfit}{bold}{\encodingdefault}{\sfdefault}{bx}{n}

\def\gD{{\mathcal{D}}}

\def\gX{{\mathcal{X}}}
\def\gY{{\mathcal{Y}}}

\DeclareMathOperator*{\argmin}{arg\,min}

\usepackage{amsmath,amsfonts,amssymb}
\usepackage{bm}
\usepackage{bbm}
\usepackage{hyperref}
\usepackage{url}
\usepackage{booktabs} 
\usepackage{multirow}
\usepackage{graphicx}
\usepackage{caption}
\usepackage{subcaption}
\usepackage{tikz}
\usepackage{algorithm}
\usepackage{algpseudocode}
\usepackage{pifont}
\usepackage{wrapfig} 
\definecolor{forestGreen}{RGB}{34, 139, 34}
\definecolor{firebrick}{RGB}{178, 34, 34}
\newcommand{\method}{{SiM}}
\newcommand{\myxmark}{\color{forestGreen}{×}}
\newcommand{\mycheckmark}{\color{firebrick}{\checkmark}}
\newcommand{\myxmarkrev}{\color{firebrick}{×}}
\newcommand{\mycheckmarkrev}{\color{forestGreen}{\checkmark}}
\usepackage[table]{xcolor}

\usepackage[accsupp]{axessibility}  

\usepackage{hyperref}
\hypersetup{hidelinks}
\hypersetup{
    colorlinks=true,
    citecolor=ForestGreen,  
    linkcolor=BrickRed,   
    urlcolor=magenta
}
\usepackage{orcidlink}

\begin{document}
\title{Training-free Task Classification\\
for Multi-Task Model Merging}
\authorrunning{J. Son et al.}

\author{
Jungyong Son\inst{1} \and
Jinwook Jung\inst{1} \and
Sungyong Baik\inst{1,2}\thanks{Corresponding author.}
}

\institute{
$^1$Department of Artificial Intelligence, 
$^2$Department of Data Science\\
Hanyang University, Republic of Korea\\
\email{\{jungyongs,jjw970517,dsybaik\}@hanyang.ac.kr}
}
  
\maketitle

\begin{abstract}
Ever since the advent of foundation models and the pretraining–finetuning paradigm, there have been numerous efforts to merge multiple task-specific experts into a single multi-task model. 
Prior work largely focuses on finding a single merged model, but it often underperforms individual experts due to parameter interference. 
To resolve this, dynamic model merging employs routing to activate task-relevant parameters per input.
However, existing routers typically require either additional training with abundant labeled datasets or assume the access to task IDs of each input at inference time.
In this work, we aim to close the gap to expert performance without additional training or task-ID-access assumption.
To this end, we formulate routing as training-free task classification for each test input.
Using singular value decomposition (SVD)-based low-rank manifold approximations for each task, \method{} scores tasks by the projection residual of the test input feature onto each task manifold and routes accordingly. 
The task manifolds are pre-computable offline from a pretrained backbone using a small per-task support set (e.g., 32 examples per task) prior to merging process, requiring no router training and no data during the merging process. 
Moreover, \method{} integrates seamlessly with subspace-/mask-based merging that represents task-expert via lightweight compressed task vectors, avoiding the need to store full expert parameters. 
Experiments across computer vision and natural language processing benchmarks under task-unknown inference demonstrate that \method{} substantially improves merged-model performance and consistently narrows the gap to individual task experts.
Our code is available at \url{https://github.com/BAIKLAB/SiM}

  
\end{abstract}  
\section{Introduction}
\begin{table}[t!]
    \caption{
        \textbf{Dynamic model merging methods and their requirements.} 
        Given \( T \) task-specific models, \( N \) denotes the number of test samples, $P_f$ denotes the number of fine-tuned model parameters, $P_c$ denotes the number of compressed fine-tuned model parameters (${P_c}\ll{P_f}$), $P_r$ denotes the number of router parameters, $M$ denotes the task-specific binary mask, $B$ denotes the number of layer blocks, $d$ denotes the feature dimension, $k$ denotes the number of ranks.
        }
    \centering
    \resizebox{0.9\linewidth}{!}{
    \begin{tabular}{lcccccc}
        \toprule
        \multirow{2}{*}{Method} & Additional & Forward & Additional &Task ID \\ 
        & training & passes & memory & requirement \\ 
        \midrule
        EMR-Merging~\cite{huang2024emr} & \myxmark & $\mathcal{O}(1)$ & $\mathcal{O}(P_f+TM)$ & \mycheckmark \\
        TALL Mask~\cite{wang2024localizing} & \myxmark & $\mathcal{O}(1)$ & $\mathcal{O}(P_f+TM)$ & \mycheckmark \\
        TWIN-Merging~\cite{lu2024twin} & \mycheckmark & $\mathcal{O}(1)$ & $\mathcal{O}(P_r+TP_c)$   & \myxmark \\
        DaWin~\cite{oh2025dawin} & \myxmark & $\mathcal{O}(N+T)$& $\mathcal{O}(TP_f)$    & \myxmark \\
        WEMoE~\cite{tang2024merging}  & \mycheckmark & $\mathcal{O}(1)$& $\mathcal{O}(BP_r+TP_f)$   & \myxmark \\
        MoW-Merging~\cite{ye2025dynamic}  & \mycheckmark & $\mathcal{O}(1)$& $\mathcal{O}(P_r+TP_f)$  & \myxmark \\
        \midrule
        \rowcolor[HTML]{FFF2CC}
        \textbf{\method (Ours)} & \myxmark & $\mathcal{O}(1)$& $\mathcal{O}(Tdk+TP_c)$  & \myxmark\\
        \bottomrule
        
    \end{tabular}}
    \label{tab:requirements}
\end{table}
With the emergence of foundation models~\cite{radford2021llearning,achiam2023gpt,touvron2023llama}, the pretraining-finetuning paradigm---the practice of pre-training foundation models followed by task-specific fine-tuning---has become one of the dominant approaches in many areas of machine learning~\cite{brown2020language,radford2019language,wei2022finetuned,wortsman2022robust}. 
The pretraining-finetuning strategy has led to the proliferation of these task-specific models, presenting new challenges related to knowledge and model management.

Accordingly, several works have attempted to integrate the knowledge from these task-specific models (or task experts) by merging them via interpolation on task-experts parameters~\cite{ilharco2023editing,matena2022merging,regmean,yadav2023ties,gargiulo2024task,iso2024daniel}.
By doing so, a merged model can replace all task experts, removing the need to store an expert model for each task and greatly reducing memory footprints.
However, these works primarily focus on finding a single static merged model, often failing to reach the performance of individual experts because of parameter interference.

In light of the challenge, few recent works have attempted to dynamically merge model parameters, employing input-adaptive interpolation of parameters~\cite{huang2024emr, wang2024localizing, gargiulo2024task, tang2024merging, lu2024twin, ye2025dynamic, oh2025dawin}.
While such input-adaptive interpolation methods have led to substantial performance improvement, few early dynamic model merging works have employed full parameters of each expert~\cite{oh2025dawin,tang2024merging,ye2025dynamic}, undermining the very goal of model merging and incurring memory overhead.
Recent works have attempted to seek a middle ground between performance and memory efficiency by compressing task vectors to efficiently store task expert knowledge~\cite{huang2024emr, wang2024localizing, gargiulo2024task,lu2024twin}.
However, they either require additional training with labeled task datasets~\cite{lu2024twin,tang2024merging,ye2025dynamic} or also require access to task identities of each input sample during inference~\cite{huang2024emr, wang2024localizing, gargiulo2024task}, which limits their applicability to real-world scenarios.

In this work, we introduce a plug-and-play dynamic model merging framework that does not require access to task identity or additional training for routing mechanism.
Particularly, we formulate the goal of finding input-adaptive task coefficients as task classification for each input data.
To this end, we propose to leverage singular value decomposition (SVD) to construct low-rank manifold approximations for each task.
Specifically, the task identity of each test input is determined by the projection residual of the test input feature onto each manifold. 
A key advantage of this approach is that task manifolds can be pre-computed offline, prior to merging process, using only a small support set (e.g., 32 examples per task) with a pre-trained backbone. 
Thus, \method~removes the need for router training or the presence of data during the actual merging phase or inference (\cref{tab:requirements}~ explicitly details the comparison between \method~and other dynamic model merging methods).
Ephemeral access to a small support set by our method mitigates issues regarding data storage requirements of previous works~\cite{ilharco2023editing, yadav2023ties}, such as memory overhead and privacy concerns.

We further note that our proposed \textbf{Si}ngular-vector-based \textbf{M}anifold (\textbf{\method}) framework can be readily integrated with existing subspace-based merging methods that compress task vectors or expert parameters~\cite{huang2024emr, wang2024localizing, gargiulo2024task}.
Notably, the integration with such lightweight compressed task vectors allows \method{} to strike a better balance in performance-memory-efficiency trade-off.
With compressed task vectors and SiM together, dynamic model merging is performed as follows: (1) task classification; (2) selecting a compressed task vector corresponding to predicted task ID; (3) adding the selected compressed task vector to pretrained model to obtain a model to use for inference.

Extensive experimental results across vision and NLP domains demonstrate that our proposed framework SiM substantially outperforms previous dynamic model merging methods that either rely on additional training or known-task-identity assumption, while overcoming these two limitations. 

\section{Related work}
\begin{figure}[t]
    \centering
    \includegraphics[width=1\textwidth]{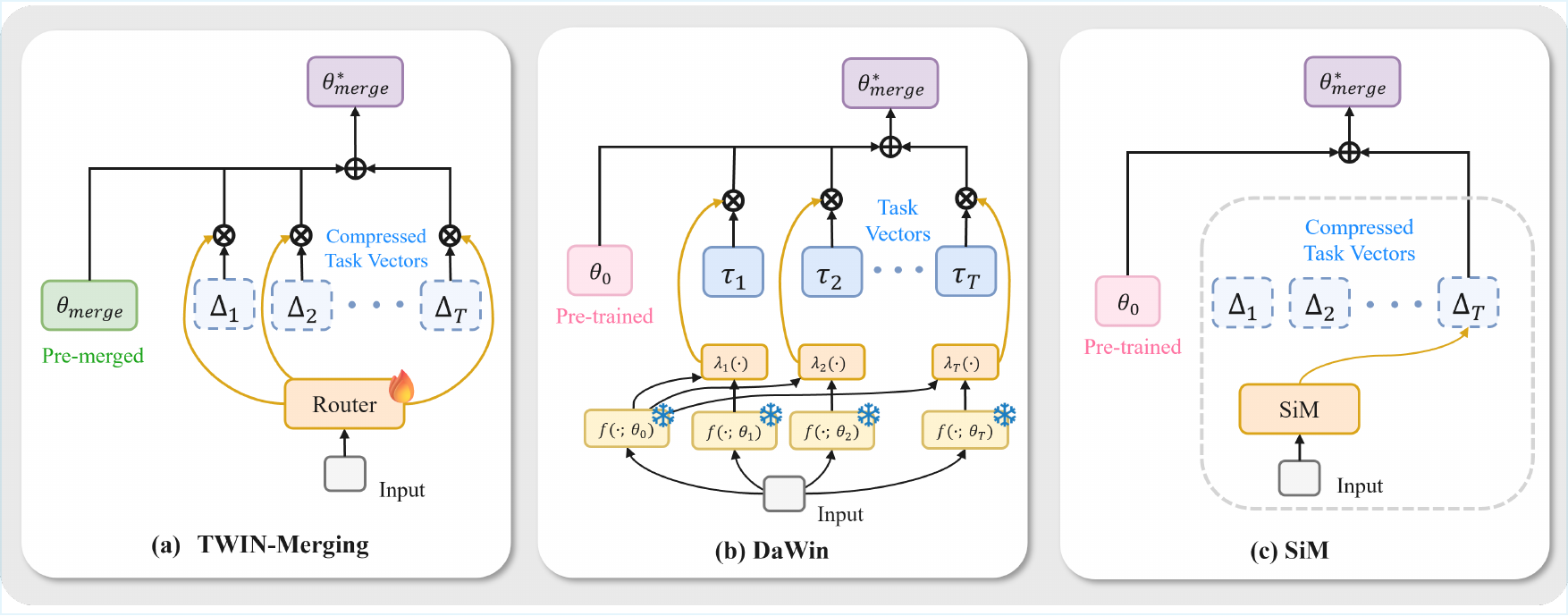}
    \caption{
        \textbf{Comparison of dynamic model merging methods.}
        Subfigure (a) shows that in TWIN-Merging, the coefficients are provided by a router that requires additional training.
        Subfigure (b) illustrates that DaWin, while training-free, requires maintaining all task-specific experts in memory to calculate coefficients.
        Subfigure (c) shows that \method~dynamically determines coefficients without any additional training and eliminates the need to hold multiple experts in memory.}
\label{fig:methods}
\end{figure}

The purpose of multi-task model merging is to build a multi-task model by combining the parameters of task-specific models. 
Task Arithmetic~\cite{ilharco2023editing} merges models by task vectors, which represent the difference between the parameters of the task-specific models and the pre-trained model. 
TIES-Merging~\cite{yadav2023ties} utilizes parameter pruning and sign election to generate task-specific masks to adjust the parameter importance, while DARE~\cite{dare} employs random dropping to reduce redundancy in task vectors. 
Additionally, methods such as RegMean~\cite{regmean} and Fisher-Merging~\cite{matena2022merging} leverage statistical information to determine optimal static weights.
More recently, TSV-M~\cite{gargiulo2024task} addresses task interference by applying whitening transformations to task-specific singular vectors, ensuring balanced joint performance. 
Similarly, Iso-C~\cite{iso2024daniel} aims to equalize the task arithmetic spectrum by scaling singular values toward their average to promote a more isotropic and balanced representation.
Alongside these approaches, recent works have explored parameter-space subspaces to refine task vectors.
DOGE~\cite{reb1} preserves shared knowledge by constraining gradient updates orthogonal to a shared parameter subspace, while Subspace Boosting~\cite{reb2} mitigates rank collapse by explicitly amplifying smaller singular values of task vectors to recover task-specific information.
Alternatively, parameter interference can also be mitigated directly during the fine-tuning stage prior to merging~\cite{miti,ortiz2023task,eff}.
However, these works are limited by their focus on a single static merged model, which frequently falls short of the performance achieved by individual experts.

Moving beyond static approaches, dynamic model merging has been introduced, an approach characterized by its adjustment of parameter importance based on test samples~\cite{lu2024twin, oh2025dawin, tang2024merging, ye2025dynamic,huang2024emr,wang2024localizing,gargiulo2024task}.
TWIN-Merging~\cite{lu2024twin} employs a router to determine parameter importance for a shared expert, but this mechanism introduces additional training overhead for the router, limiting its efficiency.
DaWin~\cite{oh2025dawin} calculates parameter importance using the Shannon entropy of the pre-trained and task-specific models, it faces scalability issues due to the requirement of multiple forward passes for entropy calculation.
WEMoE~\cite{tang2024merging} utilizes mixture-of-experts module to combine shared and task-specific experts per input, alleviating parameter interference, however at the cost of training both the router and additional experts and increasing the memory footprint.
MoW-Merging~\cite{ye2025dynamic} introduces a lightweight gating network to produce sample-wise task probabilities used as merging coefficients. However, it still requires additional training for gating network.
Apart from these architectural or training-based strategies, other efforts~\cite{huang2024emr,wang2024localizing,gargiulo2024task} have explored storing task expert knowledge more efficiently. However, these methods often necessitate access to the task identity of each input sample during inference, which restricts their applicability in real-world scenarios.

\begin{figure}[t]
    \centering
    \includegraphics[width=0.8\textwidth]{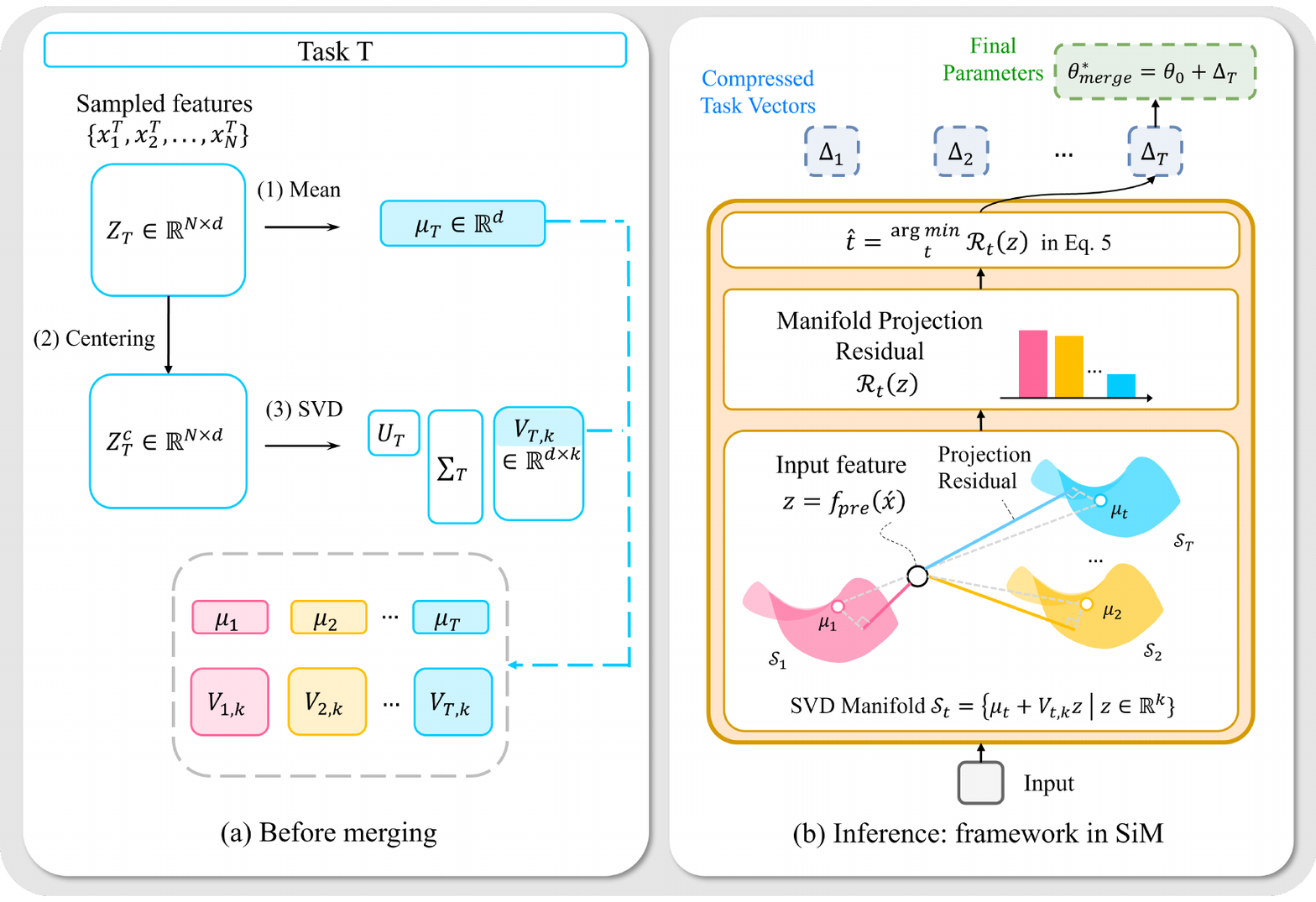}
    \caption{
        \textbf{Overview of \method.} The figure provides a detailed illustration of the \method~framework, which outlines the internal processes within the gray dashed border box shown in \cref{fig:methods}~(c). Our proposed method \method~performs task classification based on projection residual to predict task ID and selects the corresponding compressed task vector for the predicted task ID.
    }
    \label{fig:framework}
\end{figure}

To address these limitations, we introduce \method{}.
Unlike previous subspace-based methods~\cite{reb1,reb2} that exploits parameter-space subspaces to align or reweight task-vector directions for merging, \method{} constructs feature-space task subspaces for input-wise task classification and routing via projection residuals.
While ReTeX~\cite{retex} successfully repurposes this feature-space identification to train a parameter recovery module, further validating the broad flexibility of this strategy, \method{} applies these subspaces directly to dynamic model merging.
Consequently, our approach operates without the need for router training and does not require access to task identities during inference.
~\cref{fig:methods} highlights the inherent efficiency of our proposed \method~when contrasted with dynamic model merging approaches.
Furthermore, a detailed overview of the \method{} framework is illustrated in~\cref{fig:framework}.


\section{Background}
\noindent\textbf{Problem setting.}
Under pretraining-finetuning paradigm, Let $f:\gX\times\Theta\to\gY$ be a pre-trained model with its parameters $\vtheta_0\in\Theta$.
For every downstream task  $t\in\{1,...,T\}$, $f_{\vtheta_0}$ is fine-tuned on the task-specific dataset $\gD^{(t)} = \{(\vx^{(t)}_i, \allowbreak  y^{(t)}_i)\}^{N_t}_{i=1}$, consisting of input samples $\vx^{(t)}_i\in X^{(t)} \subseteq\gX$ with the corresponding labels $y^{(t)}_i\in Y^{(t)}\subseteq\gY$. 
Several early works~\cite{ilharco2023editing,matena2022merging} have focused on consolidating the knowledge of fine-tuned task experts into a single merged model via weighted combination: $\bm{\theta}=\sum_{t=1}^T \lambda_t\bm{\theta}_t$.
However, a resulting single model often fails to match the performance of each expert, due to parameter interference between experts during merging process.

\noindent\textbf{Dynamic model merging.}
To address parameter interference, dynamic model merging methods have focused on input-adaptive combination~\cite{lu2024twin,oh2025dawin,ye2025dynamic,tang2024merging}:
\begin{equation}
    \label{eq:dynamic_merging}
    \vtheta =  \sum_{t=1}^T \lambda_t(\vx)\bm{\theta}_t,  \quad \text{or} \quad \vtheta =  \vtheta_0 + \sum_{t=1}^T \lambda_t(\vx)\bm{\tau}_t,
\end{equation}    
where $\bm{\tau}_t$ is a task vector~\cite{ilharco2023editing} that represents task-specific knowledge, isolated from prior knowledge by subtracting the pre-trained model parameters from each task-expert parameters: $\bm{\tau}_t = \vtheta_t - \vtheta_0$.
However, several methods incur significant memory overhead because all full-scale expert parameters must be concurrently retained in memory during inference~\cite{oh2025dawin,tang2024merging,ye2025dynamic}.
Furthermore, other dynamic model merging methods often rely on a routing mechanism that necessitates additional training with large labeled datasets~\cite{lu2024twin,tang2024merging,ye2025dynamic}.
\section{Proposed method}
\subsection{Motivation}
\begin{figure}[t]
\begin{minipage}[t]{0.48\textwidth}
  \centering
  \includegraphics[width=0.8\linewidth]{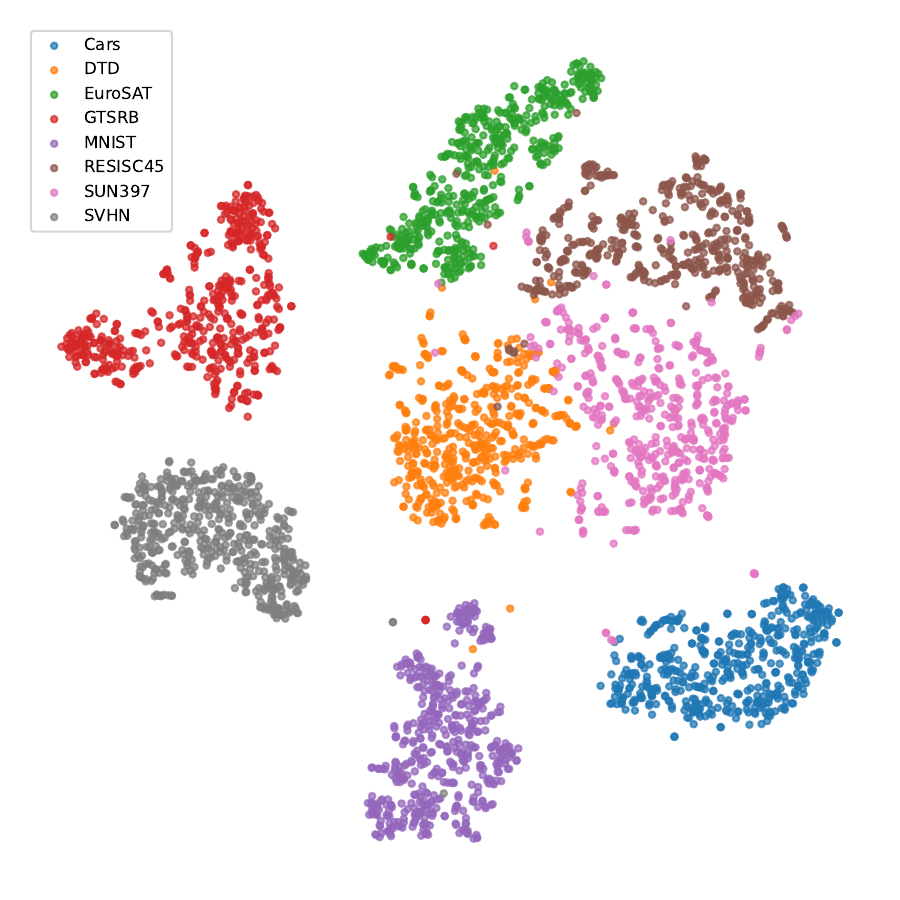}
  \caption{\textbf{Task-specific feature distributions from the CLIP ViT-B/32 across 8 vision tasks}}
  \label{fig:feature}
\end{minipage}\hfill
\begin{minipage}[t]{0.48\textwidth}
  \includegraphics[width=\linewidth]{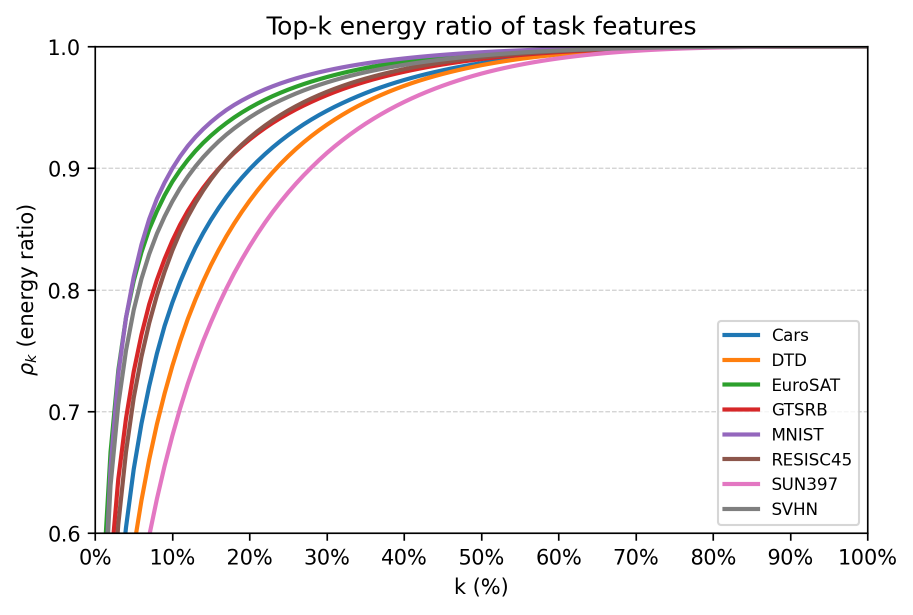}
  \caption{\textbf{Top–$k$ energy ratio of feature representations using the CLIP ViT-B/32 on 8 vision tasks}}
  \label{fig:singular}
\end{minipage}
\end{figure}

In this work, we draw inspiration from classical subspace identification~\cite{face} to address task classification challenges within the context of model merging.
This approach is predicated on the inherent structure of the feature space, where individual tasks constitute distinct and separable manifolds.
Specifically, due to inherent dataset biases, latent representations often retain dataset-specific signals, enabling the identification of the source dataset of a sample from its features~\cite{bias, battle}. 
Building upon this, we further show that feature distributions extracted from a pre-trained model are well-separated according to tasks, as illustrated in \cref{fig:feature}.



Furthermore, previous works~\cite{rankdim,intrinsic} have demonstrated that features extracted from deep neural networks inherently exhibit low-rank manifold in nature. 
To further support this observation, we conduct spectral analysis on task-specific manifold to quantify their low-rank structure. 
Specifically, singular values $\sigma_{1}\ge\sigma_{2}\ge\cdots\ge\sigma_{d}$ are extracted via Singular Value Decomposition (SVD) of task-specific feature matrix (detailed in ~\cref{estimation}). 
We then define the cumulative energy ratio as $\rho_{k}$, which represents the proportion of the total Frobenius energy captured by the top-${k}$ principal components, as follows:
\begin{equation}
\rho_k = \frac{\sum_{i=1}^k \sigma_i^2}{\sum_{j=1}^d \sigma_j^2},
\end{equation}
where $d$ denotes the feature dimension. As illustrated in \cref{fig:singular}, we observe that the cumulative energy ratio $\rho_{k}$ reveals that the top 10\% of the singular values already capture more than 80\% of the Frobenius energy across tasks.
This high concentration of energy indicates that task-specific features effectively reside on low-dimensional subspaces rather than being uniformly distributed.
Consequently, this observation demonstrates the feasibility of a subspace-based approach for task classification.

\subsection{Task manifold construction}
\label{estimation}
To characterize the distribution for each task, we construct a task-specific subspace using representations extracted from the pre-trained model.
For each task $t\in\{1,...,T\}$, we first compute the empirical mean $\boldsymbol{\mu}_t\in \mathbb{R}^{d}$, using \( N \) randomly gathered samples \( \{{\vx}^{(t)}_1, {\vx}^{(t)}_2, \ldots, {\vx}^{(t)}_N\} \subset \mathcal{X} \) of the training set for task \( t \):

\begin{equation}
{\boldsymbol{\mu}}_t = \frac{1}{N} \sum_{i=1}^N f_{\text{pre}}({\vx}^{(t)}_i) .
\label{eq:mean}
\end{equation}
To identify the principal structure of the task, we construct a centered
feature matrix ${\mathbf{Z}}_t^c\in\mathbb{R}^{N \times d}$ by subtracting the empirical mean $\boldsymbol{\mu}_t$ from each extracted feature vector ${\vz}_t\in \mathbb{R}^{d}$. We then perform Singular Value Decomposition (SVD) on the centered matrix ${\mathbf{Z}}_t^c$ to obtain the right-singular vectors:
\begin{equation}
{\mathbf{Z}}_t^c = \mathbf{U}_t \mathbf{\Sigma}_t \mathbf{V}_t^\top,
\label{eq:singular}
\end{equation}
where the columns of $\mathbf{V}_t\in\mathbb{R}^{d \times d}$ represent the principal components of the feature distribution for task $t$. 
Then, we retain the top-$k$ basis vectors to construct the subspace matrix $\mathbf{V}_{t,k} \in \mathbb{R}^{d \times k}$ with $k\ll d$.
Consequently, for each task $t$, we store only the mean $\boldsymbol{\mu}_t$ and the subspace matrix $\mathbf{V}_{t,k}$ as the essential task signatures required for the subsequent inference stage. 
Notably, task manifolds can be pre-computed from a pre-trained backbone using a small set per task, eliminating the need for additional training or data storage during the merging process.
Unless otherwise indicated, a sample size of \( N = 32 \) and  $k$ is set to 10\% of the feature dimension.
We provide extensive ablation studies justifying these choices, demonstrating robustness of \method{} to varying rank $k$, sample sizes $N$, and even support-set selection strategies in Appendix~\cref{app:ab_rank,app:ab_seed,app:support-set}.

\subsection{Inference}
\noindent\textbf{Task classification via manifold projection.}
Given a test sample \( \acute\vx \in \mathcal{D}_\text{test} \subset \mathcal{X} \) and its feature vector ${\vz} = f_{\text{pre}}({\acute\vx}) \in\mathbb{R}^{d}$,
we define the manifold projection residual of ${\vz}$ with respect to task $t$ as follows:
\begin{equation}
\mathcal{R}_t(\mathbf{z}) = || (\mathbf{I} - \mathbf{V}_{t,k} \mathbf{V}_{t,k}^\top) ({\vz} - \boldsymbol{\mu}_t) ||_2,
\label{eq:residual}
\end{equation}
where $\mathbf{I}\in\mathbb{R}^{d \times d}$ denotes the identity matrix, $\boldsymbol{\mu}_t$ and $\mathbf{V}_{t,k}$ are the empirical mean and the basis matrix of the subspace for task $t$, respectively.
The predicted task ID $\hat{t}$ for each sample $\acute\vx$ is determined by selecting the task that has the smallest projection residual:
\begin{equation}
\hat{t} = \argmin_t \mathcal{R}_t({\vz}).
\label{eq:prediction}
\end{equation}
Intuitively, each sample is assigned to the task whose manifold aligns with the sample most closely in the feature space.


\noindent\textbf{Group batch inference.}
During inference, we process a batch of $B$ samples to leverage computational efficiency. After determining the predicted task IDs $\hat{t}$ for all samples in the batch, we group them according to their predicted tasks. 
For each task, we select the corresponding samples \( \acute\vx_{\hat{t}}  \) and task vectors $\bm{\tau_{\hat{t}}}$.
Then, batch inference is conducted on the selected samples with a resulting merged parameters $\vtheta^\star_\text{merge} = \vtheta_0 + \bm{\tau}_{\hat{t}}$.

To achieve strong performance and efficiency, we integrate our method \method{} with several subspace-based model merging methods that replace task vectors $\bm{\tau}_t$ with efficiently compressed task vectors $\bm\Delta_{\text{t}}$.

\noindent EMR-Merging (EMR)~\cite{huang2024emr}: 
EMR first aggregates task vectors into a single pre-merged vector $\bm\tau_{\text{merge}}$ using a sign-and-magnitude rule. Then, for each task $t$, it derives a task-specific binary mask $\mathcal{M}_t=\mathbbm{1}\{\bm\tau_t\odot\bm\tau_{\text{merge}}>0\}$ and a rescaling factor $\lambda_t$.
    The masked task vector for task $t$ is obtained as  $\bm\Delta_{\text{t}} = \lambda_t\,(\mathcal{M}_t\odot \bm\tau_{\text{merge}})$.

\noindent TALL-Mask + Task Arithmetic (TM-TA)~\cite{wang2024localizing}: 
     This method first computes a merged task vector \( \bm\tau_\text{merge} = \alpha\sum_{t=1}^T \bm\tau_t \) using Task Arithmetic~\cite{ilharco2023editing}.
     Then, for each task $t$, it computes a binary mask
    $\mathcal{M}_t=\mathbbm{1}\{|\bm\tau_t|\ge |\bm\tau_{\text{merge}}-\bm\tau_t|\cdot \lambda_t\}$. The hyperparameter \( \lambda_t \) determines how much information to extract from the merged task vector.
    The masked task vector update for task $t$ is $\bm\Delta_{\text{t}} =\mathcal{M}_t\odot \bm\tau_{\text{merge}}$.

\noindent TALL-Mask + TIES-Merging (TM-TIES)~\cite{wang2024localizing}: 
    This method is one of the variants of TALL-Mask.
    Instead of computing the merged task vector \( \bm\tau_\text{merge} \) using Task Arithmetic, this variant leverages TIES-Merging for its calculation~\cite{yadav2023ties}.
    The remaining steps are identical to those previously explained.

\noindent TSV-C~\cite{gargiulo2024task}: 
    TSV-C represents task vectors as layer-wise task matrices and applies singular value decomposition (SVD).
    For each task $t$, only the most top-$k$ significant singular vectors are retained, yielding a compressed task vector $\bm\Delta_{\text{t}} = \sum_{i=1}^{k}\mathbf{U}_{t,i}\mathbf{\Sigma}_{t,i}\mathbf{V}_{t,i}^\top$. 

The plug-and-play design nature of \method{} allows us to empower efficient subspace-based model merging methods with the ability to perform dynamic merging for task-agnostic scenarios, where task ID of each input is unknown. 
Furthermore, integration with subspace-based model merging methods creates synergy as their compressed task vectors enable memory-efficient inference.
A detailed description of the complete \method{} procedure, along with a detailed comparison of ephemeral data usage of \method{} against standard routing pipelines, is provided in Appendix~\cref{app:implemantation}.

\begin{figure}[t]
\begin{minipage}[t]{0.46\textwidth}
  \centering
  \includegraphics[width=\linewidth]{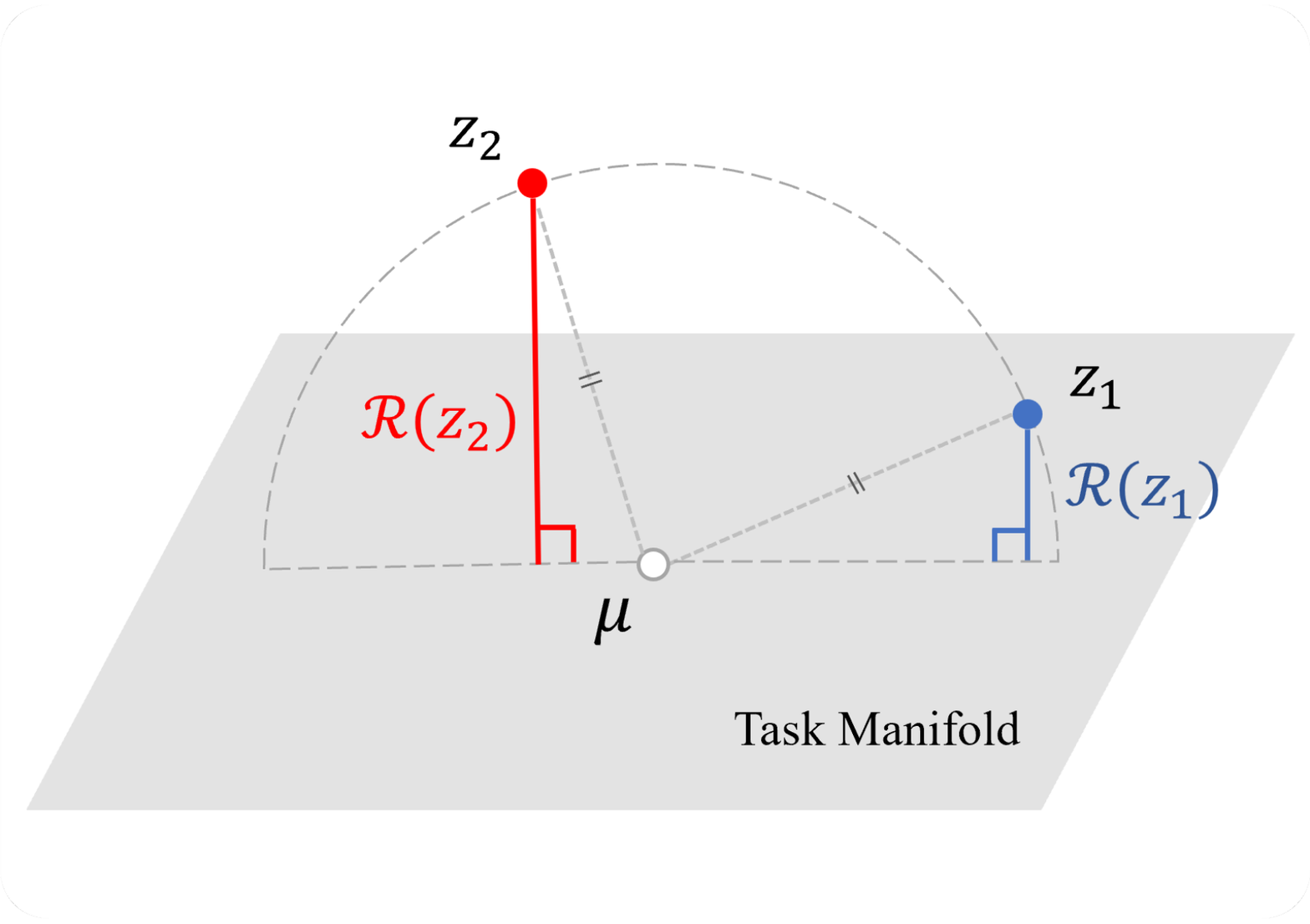}
  \caption{
  \textbf{Manifold projection residual vs. Euclidean distance.}
  Two features ${{\vz}_1}$ and ${{\vz}_2}$ have the same Euclidean distance to the task mean $\boldsymbol{\mu}$, but their projection residuals differ depending on their alignment with the task manifold.}
  \label{fig:residual}
\end{minipage}\hfill
\begin{minipage}[t]{0.48\textwidth}
  \includegraphics[width=\linewidth]{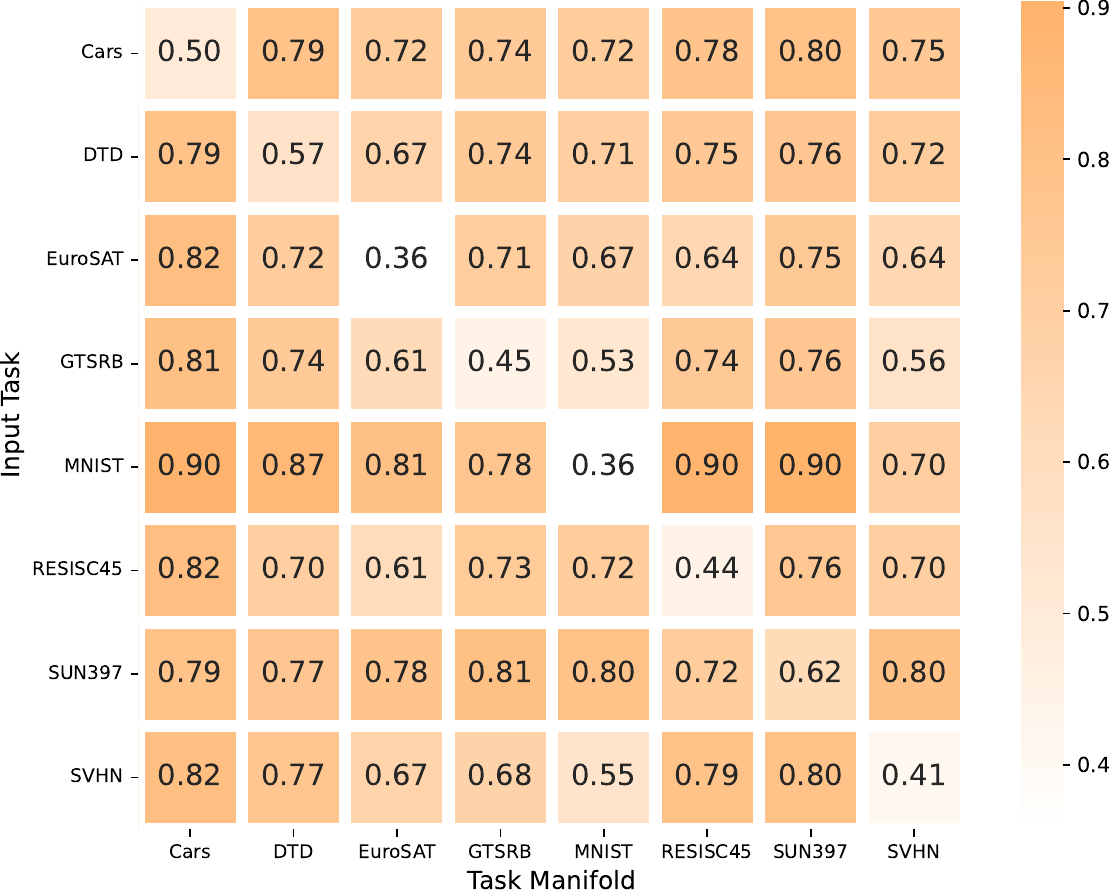}
  \caption{\textbf{Average residual ratio $\lambda_{z,t}$ across tasks and subspaces on 8 vision tasks using the CLIP ViT-B/32}}
  \label{fig:ratio}
\end{minipage}
\end{figure}

\subsection{Analysis of projection residual}
\label{sec:theoretical analysis}
\textcolor{black}{
In this section, we justify the manifold projection residual formulation for task classification, in contrast to simple distance-based metric, such as Euclidean distance.
Although task feature distributions exhibit inherent separability (shown in \cref{fig:feature}), simple distance-based metrics (e.g., Euclidean distance) provide only a coarse signal that neglects the intrinsic low-dimensional structure of each task.
}
We therefore investigate how the projection residual operator $(\mathbf{I} - \mathbf{V}_{t,k} \mathbf{V}_{t,k}^\top)$ can leverage this geometric orientation to significantly enhance feature discriminability, enabling more robust task classification.

The projection residual operator removes the component of the mean-centered feature aligned with the task subspace and retains only the orthogonal component.
Consequently, even when two features have same Euclidean distances to the task mean, the residual magnitude is small if the feature direction aligns with the task subspace and large otherwise, as illustrated in~\cref{fig:residual}.

To further validate the manifold projection residual formulation for task classification, we analyze how well the projection residual leads to the selection of appropriate task. 
To quantify the relative residual magnitude induced by each task-specific subspace in a scale-invariant manner, we define the residual ratio $\lambda_{z,t}$ as follows:
\begin{equation}
\lambda_{z,t} =
\left\|(\mathbf{I}-\mathbf{V}_{t,k}\mathbf{V}_{t,k}^\top)
\frac{{\vz}-\boldsymbol{\mu}_t}{\|{\vz}-\boldsymbol{\mu}_t\|_2}
\right\|_2.
\label{eq:ratio}
\end{equation}
A smaller $\lambda_{z,t}$ indicates stronger alignment between the input and the subspace of task $t$, while a larger value indicates weaker alignment. 
Comparing $\lambda_{z,t}$ across subspaces therefore yields a discriminative gap in residual magnitudes.
\cref{fig:ratio} illustrates the average $\lambda_{z,t}$ computed for each input-task/subspace pair across 8 vision tasks.
The result shows that the diagonal entries are consistently the lowest, indicating that inputs from each task are most aligned with their own task subspace.
As a result, for a given input, the projection residual operator yields a much smaller residual under its own task subspace and larger residuals under other task subspaces.
This gap in residual ratio indicates that task-specific subspaces provide a discriminative signal, improving separability in residual space and facilitating reliable task identification. 
This corresponds to the superior task classification performance (more details in \cref{sec:task_cls}), validating the effectiveness of our proposed method \method{}.

\begin{table*}[t]
\caption{\textbf{Multi-task performance of merged CLIP ViT models for computer vision tasks across different number of tasks}. We report experimental results for ViT-B/32, ViT-B/16, and ViT-L/14 on 8, 14, and 20 vision tasks. Bold values represent the best performance among all methods, excluding the individual task-specific baselines.
}
\centering
\setlength\tabcolsep{2.pt}
\renewcommand{\arraystretch}{1.2}
\footnotesize
\resizebox{\textwidth}{!}{
\begin{tabular}{l|ccc|ccc|ccc}
\toprule
\multicolumn{1}{l|}{\multirow{2}{*}{\textbf{Method}}} & \multicolumn{3}{c|}{ViT-B/32} & \multicolumn{3}{c|}{ViT-B/16} & \multicolumn{3}{c}{ViT-L/14}\\
\cmidrule{2-10}
& {8 tasks} & {14 tasks} & {20 tasks} & {8 tasks} & {14 tasks} & {20 tasks} & {8 tasks} & {14 tasks}  & {20 tasks}\\ 
\midrule
\multicolumn{1}{l|}{Fine-tuned} & 92.8 & 90.9 & 91.3 & 94.7 & 92.8 & 92.8 & 95.9 & 94.3 & 94.8 \\
\midrule
\rowcolor[HTML]{EFEFEF}
        \multicolumn{10}{l}{\textbf{\textit{(Static model merging)}}} \\
\multicolumn{1}{l|}{Task Arithmetic~\cite{ilharco2023editing}} & 70.8 & 65.3 & 60.5 & 75.4 & 70.5 & 65.8 & 84.9 & 79.4 & 74.0 \\
\multicolumn{1}{l|}{TIES-Merging~\cite{yadav2023ties}} & 75.1 & 68.0 & 63.4 & 79.7 & 73.2 & 68.2 & 86.9 & 79.5 & 75.7\\
\multicolumn{1}{l|}{TSV-M~\cite{gargiulo2024task}} & 85.7 & 80.1 & 77.1 & 89.0 & 84.6 & 80.6 & 93.0 & 89.2 & 87.7\\
\multicolumn{1}{l|}{Iso-C~\cite{iso2024daniel}} & 84.7 & 79.9  & 74.6  & 89.2 & 84.5 & 79.1 & 93.3 & 89.4 & 87.7\\
\midrule
\rowcolor[HTML]{EFEFEF}
        \multicolumn{10}{l}{\textbf{\textit{(Dynamic model merging)}}} \\
\multicolumn{1}{l|}{TWIN-Merging~\cite{lu2024twin}} & 84.0 & 70.0 & 57.5 & 91.4 & 78.4 & 63.1 & 93.7 & 86.2 & 74.8 \\
\multicolumn{1}{l|}{DaWin~\cite{oh2025dawin}} & 89.0 & 73.8 & 52.8 & 87.1 & 77.8 & 62.8 & 91.6 & 82.6 & 77.5 \\
\multicolumn{1}{l|}{WEMoE~\cite{tang2024merging}} & 90.4 & 83.1& 74.4 & 93.1 & 84.0 &  76.6 &  94.8 &  87.0 &  75.7 \\
\multicolumn{1}{l|}{MoW-Merging~\cite{ye2025dynamic}} &  88.1 &  83.2 &  79.3 &  93.7 &  79.3 &  78.2 &  94.9 &  78.8 &  81.8 \\
\midrule
\rowcolor[HTML]{FFF2CC}
\multicolumn{1}{l|}{\textbf{{EMR~\cite{huang2024emr} + \method~(Ours)}}} & 90.7 & 87.0 & 86.1 & 92.9 & 90.0 & 89.1 & 95.0 & 92.5 & 92.2 \\
\rowcolor[HTML]{FFF2CC}
\multicolumn{1}{l|}{\textbf{{TM-TA~\cite{wang2024localizing} + \method~(Ours)}}} & 92.3 & \textbf{89.5} & \textbf{89.6} & 94.0 & 91.2 & \textbf{91.5} & 92.0 & 89.1 & 89.6 \\
\rowcolor[HTML]{FFF2CC}
\multicolumn{1}{l|}{\textbf{{TM-TIES~\cite{wang2024localizing} + \method~(Ours)}}} & \textbf{92.7} & 88.5 & 82.2 & \textbf{94.3} & 90.5 & 84.5 & 95.0 & 91.8 & 91.4 \\
\rowcolor[HTML]{FFF2CC}
\multicolumn{1}{l|}{\textbf{{TSV-C~\cite{gargiulo2024task} + \method~(Ours)}}} & 92.1 & 89.4 & 88.6 & 94.0 & \textbf{91.5} & 91.1 & \textbf{95.4} & \textbf{93.6} & \textbf{92.8} \\
\bottomrule
\end{tabular}}
\label{tab:visionacc}
\end{table*}
\section{Experiments}

\begin{table*}[t]
\caption{\textbf{Multi-task performance of merged T5-large across 7 NLP tasks.}
Bold values represent the best performance among all methods, excluding the individual task-specific baselines.
Underlines indicate the case where fine-tuning performance is surpassed.
}
\centering
\setlength\tabcolsep{4.5pt}
\renewcommand{\arraystretch}{1.2}
\footnotesize
\resizebox{\textwidth}{!}{
\begin{tabular}{l|ccccccc|c}
\toprule
        \multicolumn{1}{l|}{\textbf{Method}} & PAWS & QASC & QuaRTz & StoryCloze & WikiQA & Winogrande & WSC  & \textbf{Avg.} \\
            \midrule
            \multicolumn{1}{l|}{Fine-tuned} & 94.4 & 98.9 & 87.8 & 90.8 & 96.0 & 74.7 & 79.2 & 88.8\\
            \midrule
            \rowcolor[HTML]{EFEFEF}
            \multicolumn{9}{l}{\textbf{\textit{(Static model merging)}}} \\
            \multicolumn{1}{l|}{Task Arithmetic~\cite{ilharco2023editing}} & 77.8 & 96.0 & 78.6 & 86.4 & 59.1 & 62.3 & 52.8 & 73.3 \\
            \multicolumn{1}{l|}{TIES-Merging~\cite{yadav2023ties}} & 81.5 & 96.2 & 80.1 & 83.6 & 64.9 & 66.5 & 65.3 & 76.9 \\
            \multicolumn{1}{l|}{TSV-M~\cite{gargiulo2024task}} & 85.0 & 92.4 & 66.0 & 81.1 & 88.4 & \underline{87.0} & 57.0 & 79.6 \\
            \multicolumn{1}{l|}{Iso-C~\cite{iso2024daniel}} & 72.9 & 94.8 & 50.5 & 80.7 & 92.5 & \underline{75.7} & 54.0 & 74.4 \\
            \midrule
            \rowcolor[HTML]{EFEFEF}
            \multicolumn{9}{l}{\textbf{\textit{(Dynamic model merging)}}} \\
            \multicolumn{1}{l|}{TWIN-Merging~\cite{lu2024twin}} & 93.2 & 96.0 & \textbf{87.1} & 85.6 & 77.1 & 70.8 & \textbf{69.7} & 82.8  \\
            \multicolumn{1}{l|}{DaWin~\cite{oh2025dawin}} & 85.6 & 96.1 & 81.4 & 73.2 & 70.2 & 68.7 & 57.9 & 76.2 \\
            \multicolumn{1}{l|}{WEMoE~\cite{tang2024merging}} & 93.4 & 95.4 & 74.5 & 79.6 & 94.1 & \underline{90.4} & 57.0 & 83.4 \\
            \multicolumn{1}{l|}{MoW-Merging~\cite{ye2025dynamic}} & \underline{94.7} & 90.2 & 79.0 & 83.1 & 56.6 & \underline{\textbf{93.8}} & 51.0 & 78.3 \\
            \midrule
            \rowcolor[HTML]{FFF2CC}
            \multicolumn{1}{l|}{\textbf{EMR~\cite{huang2024emr} + \method}} & \underline{95.0} & 95.8 & 83.5 & 87.8 & 92.4 & \underline{80.6} & 59.0 & 84.9 \\
            \rowcolor[HTML]{FFF2CC}
            \multicolumn{1}{l|}{\textbf{TM-TA~\cite{wang2024localizing} + \method}} & 88.3 & 95.6 & 81.5 & 88.0 & \textbf{95.2} & \underline{87.2} & 53.0 & 84.1 \\
            \rowcolor[HTML]{FFF2CC}
            \multicolumn{1}{l|}{\textbf{TM-TIES~\cite{wang2024localizing} + \method}} & \underline{\textbf{95.5}} & 97.0 & 83.5 & 89.2 & 94.2 & \underline{91.1} & 49.0 & \textbf{85.6} \\
            \rowcolor[HTML]{FFF2CC}
            \multicolumn{1}{l|}{\textbf{TSV-C~\cite{gargiulo2024task} + \method}} & \underline{95.0} & \textbf{97.6} & 84.5 & \textbf{89.7} & 94.4 & 74.5 & 55.0 & 84.4 \\
        \bottomrule
\end{tabular}}
\label{tab:nlpacc}
\end{table*}

\begin{table*}[t]
\centering
\caption{
\textbf{Multi-task performance on seen versus unseen tasks under distribution shift.}
The ``seen'' split corresponds to the 8-task vision benchmark used in the main experiments, whereas the ``unseen'' split consists of the additional 6 datasets from the 14-task benchmark that are not included in the 8-task set.
Averaged performance across all 14 tasks is reported in the last column.
}

\label{tab:ood_tasks}
\setlength\tabcolsep{4.0pt}
\resizebox{\textwidth}{!}{
\begin{tabular}{l|ccc}
\toprule
\textbf{Method} 
& \textbf{Seen (8 tasks)} 
& \textbf{Unseen (6 tasks)} 
& \textbf{Overall (14 tasks)} \\
\midrule
\rowcolor[HTML]{EFEFEF}
\multicolumn{4}{l}{\textbf{\textit{(Static model merging)}}} \\
Task Arithmetic~\cite{ilharco2023editing} & 70.8 & 44.3 & 59.4 \\
TIES-Merging~\cite{yadav2023ties}    & 75.1 & 55.2 & 66.6 \\
TSV-M~\cite{gargiulo2024task}    & 85.7 & 60.5 & 74.9 \\
Iso-C~\cite{iso2024daniel}    & 84.7 & 60.2 & 74.2 \\
\midrule
\rowcolor[HTML]{EFEFEF}
\multicolumn{4}{l}{\textbf{\textit{(Dynamic model merging)}}} \\
TWIN-Merging~\cite{lu2024twin}                             & 84.0 & 62.6 & 74.8 \\
DaWin~\cite{oh2025dawin}                                    & 89.0 & 57.5 & 75.5 \\
WEMoE~\cite{tang2024merging}                             & 90.4 & 59.2 & 77.0 \\
MoW-Merging~\cite{ye2025dynamic}                             & 88.1 & 59.0 & 75.6 \\
\midrule
\rowcolor[HTML]{FFF2CC}
\textbf{EMR~\cite{huang2024emr} + \method{} (Ours) }                 & 90.7 & 57.9 & 76.6 \\
\rowcolor[HTML]{FFF2CC}
\textbf{TM-TA~\cite{wang2024localizing} + \method{} (Ours)}                        & 92.3 & 62.6 & 79.6 \\
\rowcolor[HTML]{FFF2CC}
\textbf{TM-TIES~\cite{wang2024localizing} + \method{} (Ours) }                     & \textbf{92.7} & \textbf{63.3} & \textbf{80.1} \\
\rowcolor[HTML]{FFF2CC}
\textbf{TSV-C~\cite{gargiulo2024task} + \method{} (Ours) }                     & 92.1 & 58.0 & 77.5 \\
\bottomrule
\end{tabular}
}
\end{table*}

\subsection{Setup}

\noindent\textbf{Baselines.}
We compared applying our proposed method to existing dynamic model merging methods against several baselines such as EMR-Merging ~\cite{huang2024emr}, TALL-Mask + Task Arithmetic, TALL-Mask + TIES~\cite{wang2024localizing, ilharco2023editing, yadav2023ties}, and TSV-C~\cite{gargiulo2024task}. Furthermore, we compare against other dynamic model merging methods that require additional training, including TWIN-Merging~\cite{lu2024twin}, DaWin~\cite{oh2025dawin}, WEMoE~\cite{tang2024merging} and MoW-Merging~\cite{ye2025dynamic} as well as representative static model merging baselines~\cite{ilharco2023editing, yadav2023ties, gargiulo2024task, iso2024daniel}.


\noindent\textbf{Evaluation settings.} 
\label{subsec:exp_details}
For evaluation, we follow the settings of TALL-Mask~\cite{wang2024localizing} for the computer vision tasks, and the settings of TIES-Merging~\cite{yadav2023ties} for the 7 NLP tasks.
We evaluate our method across four task scenarios. 
The 8 computer vision task scenario consists of: (1) SUN397~\cite{xiao2016sun}, (2) Cars~\cite{krause20133d}, (3) RESISC45~\cite{cheng2017remote}, (4) EuroSAT~\cite{helber2019eurosat}, (5) SVHN~\cite{netzer2011reading}, (6) GTSRB~\cite{stallkamp2011german}, (7) MNIST~\cite{deng2012mnist}, and (8) DTD~\cite{cimpoi2014describing}. 
The 14 computer vision task scenario adds: (9) CIFAR100~\cite{krizhevsky2009learning}, (10) STL10~\cite{coates2011analysis}, (11) Flowers102~\cite{nilsback2008automated}, (12) OxfordIIITPet~\cite{parkhi2012cats}, (13) PCAM~\cite{veeling2018rotation}, and (14) FER2013~\cite{goodfellow2013challenges}. 
The 20 computer vision task scenario further includes: (15) EMNIST~\cite{cohen2017emnist}, (16) CIFAR10~\cite{krizhevsky2009learning}, (17) Food101~\cite{bossard2014food}, (18) FashionMNIST~\cite{xiao2017fashion}, (19) RenderedSST2~\cite{socher2013recursive,radford2019language}, and (20) KMNIST~\cite{clanuwat2018deep}. 

The 7 NLP task scenario comprises: (1) QASC~\cite{khot2020qasc},
(2) WikiQA~\cite{yang2015wikiqa},
(3) QuaRTz~\cite{tafjord2019quartz} for question answering,
(4) PAWS~\cite{zhang2019paws} for paraphrase identification,
(5) Story Cloze~\cite{sharma2018tack}  for sentence completion,
(6) Winogrande~\cite{sakaguchi2020wino},
(7) WSC~\cite{levesque2012wsc} for coreference resolution.
We perform merging on CLIP~\cite{radford2021llearning} ViT-\{B/32, B/16, L/14\}~\cite{dosovitskiy2021an} for the computer vision tasks and on T5-large~\cite{raffel2020exploring} for the NLP tasks.
The model checkpoints used in our experiments are obtained from the following sources: TALL-Mask~\cite{wang2024localizing} (for the computer vision tasks) and TIES-Merging~\cite{yadav2023ties} (for the 7 NLP tasks).
Further experimental details are provided in Appendix Sec. \textcolor{BrickRed}{C}.

\subsection{Main results}
\noindent\textbf{Computer vision tasks.} 
Experimental results for merging ViT-B/32, ViT-B/16 and ViT-L/14 across the 8, 14, and 20 vision tasks are summarized in ~\cref{tab:visionacc}. 
These results demonstrate that our approach, integrating with subspace-based merging methods, consistently surpasses existing model merging methods across various model sizes without requiring prior knowledge of task IDs or additional training.
We also investigate the scalability of our method by evaluating its performance on task sets comprising more than 14 tasks. Our method consistently maintains the performance, highlighting the significance of our proposed methodology.
For more extreme scenarios, we provide further results in Appendix~\cref{app:larger,app:similar} demonstrating the robustness of \method{} on larger scales (up to 50 tasks) and highly ambiguous fine-grained domains (e.g., CUB-200).

\noindent\textbf{NLP tasks.} 
We further assess the effectiveness of our method by evaluating its performance on
NLP tasks, in addition to the vision tasks. The corresponding experimental results for the NLP
tasks are shown in ~\cref{tab:nlpacc}.
Our method demonstrates near fine-tuning performance even on NLP tasks. 
Notably, it surpasses fine-tuning performance on PAWS and Winogrande. 
This suggests that \method{} is capable of identifying a superior model compared to the original expert model.

\noindent\textbf{Robustness to unseen tasks.}
We evaluate task classification on seen and unseen tasks to test robustness to distributional shifts.
Concretely, we treat the 8-task vision benchmark used in the main experiments as the set of \emph{seen} tasks, and use the additional 6 datasets from the 14-task vision benchmark as \emph{unseen} tasks.
For each input, \method{} identifies the task whose subspace yields the minimum projection residual among the seen tasks and selects the corresponding compressed task vector.
As summarized in ~\cref{tab:ood_tasks}, among baselines, TWIN-Merging~\cite{lu2024twin} improves robustness to unseen tasks, but still exhibit a noticeable drop in accuracy compared to seen tasks.
In contrast, \method{} obtains the best overall performance.
These results indicate that incorporating the \method{} enables dynamic model merging methods to handle unseen tasks effectively, thereby mitigating the impact of feature distribution shift as the task set evolves.

 \begin{table}[t]
     \centering
     \caption{\textbf{Inference cost with CLIP ViT-B/32.} We report the computation costs of model merging methods across all tasks in the 8 computer vision task scenarios, measured on an NVIDIA GeForce RTX 3090.
     }
    \centering
    \setlength\tabcolsep{3.0pt}
    \resizebox{\textwidth}{!}{
     \begin{tabular}{ccccc}
         \toprule
         \multicolumn{1}{c|}{\textbf{Method}} & Batch inference& Latency (per input) & VRAM (GB) & Avg. performance \\
         \midrule
         \rowcolor[HTML]{EFEFEF}
         \multicolumn{5}{l}{\textbf{\textit{(Static model merging)}}} \\
         \multicolumn{1}{l|}{Task-Arithmetic~\cite{ilharco2023editing}} & \mycheckmarkrev & 0.0008s & 1.3 & 70.8 \\
         \multicolumn{1}{l|}{TIES-Merging~\cite{yadav2023ties}} & \mycheckmarkrev & 0.0008s & 1.3 & 75.1 \\
         \multicolumn{1}{l|}{TSV-M~\cite{gargiulo2024task}} & \mycheckmarkrev & 0.0008s & 1.3 & 85.7 \\
         \multicolumn{1}{l|}{Iso-C~\cite{iso2024daniel}} & \mycheckmarkrev & 0.0008s & 1.3 & 84.7 \\
         \midrule
         \rowcolor[HTML]{EFEFEF}
         \multicolumn{5}{l}{\textbf{\textit{(Dynamic model merging)}}} \\
         \multicolumn{1}{l|}{TWIN-Merging~\cite{lu2024twin}} & \myxmarkrev & 0.03s & 3.2 & 84.0 \\
         \multicolumn{1}{l|}{DaWin~\cite{oh2025dawin}}& \myxmarkrev & 0.63s & 5.5 & 89.0\\
         \multicolumn{1}{l|}{WEMoE~\cite{tang2024merging}}& \myxmarkrev & 0.02s & 4.3 & 90.4 \\
         \multicolumn{1}{l|}{MoW-Merging~\cite{ye2025dynamic}} & \mycheckmarkrev & 0.008s & 4.9 & 88.1 \\
         \midrule
         \rowcolor[HTML]{FFF2CC}
         \multicolumn{1}{l|}{\textbf{EMR~\cite{huang2024emr} + \method~(Ours)} } & \mycheckmarkrev & 0.004s & 1.9 & 90.7 \\
         \rowcolor[HTML]{FFF2CC}
         \multicolumn{1}{l|}{\textbf{TM-TA~\cite{wang2024localizing} + \method~(Ours)} }& \mycheckmarkrev  & 0.004s & 1.8 & 92.3 \\
         \rowcolor[HTML]{FFF2CC}
         \multicolumn{1}{l|}{\textbf{TM-TIES~\cite{wang2024localizing} + \method~(Ours)} } & \mycheckmarkrev  & 0.003s & 1.8 & 92.7 \\
         \rowcolor[HTML]{FFF2CC}
         \multicolumn{1}{l|}{\textbf{TSV-C~\cite{gargiulo2024task} + \method~(Ours)} } & \mycheckmarkrev  & 0.003s & 1.6 & 92.1 \\
         \bottomrule
     \end{tabular}}
     \label{tab:cost} 
 \end{table}

\begin{table*}[t]
    \centering
    \caption{\textbf{Training-free task classification performance comparison across different distance-based metrics.}
We compare task classification performance using Euclidean distance, cosine similarity, Mahalanobis distance, $k$-NN and our final model.
}

  \resizebox{1\textwidth}{!}{
    \begin{tabular}{l|cccccccc|c}
        \toprule
            \multicolumn{1}{c|}{\textbf{Method}} & Cars & DTD & EuroSAT & GTSRB & MNIST & RESISC45 & SUN397 & SVHN & \textbf{Avg.} \\
            \midrule
            \multicolumn{1}{c|}{Euclidean dist.} & 99.7 & 96.9 & 86.5 & 90.0 & 100 & 95.7 & 97.8 & 99.6 & 95.8 \\
            \multicolumn{1}{c|}{Cosine sim.} & 99.7 & 96.5 & 86.5 & 89.7 & 100 & 95.3 & 97.0 & 99.6 & 95.5 \\
            \multicolumn{1}{c|}{Mahalanobis dist.} & 98.6 & 92.3 & 95.4 & 97.9 & 99.1 & 91.5 & 99.8 & 97.5 & 96.5 \\
            \multicolumn{1}{c|}{$k$-NN ($k$=1)} & 99.9 & 91.8 & 99.2 & 97.1 & 99.9 & 79.5 & 75.7 & 92.9 & 92.0 \\
            \midrule
            \rowcolor[HTML]{FFF2CC}
            \multicolumn{1}{c|}{\textbf{\method}} & 99.8 & 98.6 & 99.9 & 98.4 & 100 & 98.8 & 95.4 & 99.9 & \textbf{99.0} \\
            
        \bottomrule
    \end{tabular}
    }
\label{tab:ablation} 
\end{table*}
\subsection{Computational costs}
In this section, we investigate the efficiency of \method{}, particularly in comparison to another dynamic model merging method: TWIN-Merging~\cite{lu2024twin}, DaWin~\cite{oh2025dawin}, WEMoE~\cite{tang2024merging} and MoW-Merging~\cite{ye2025dynamic}.
As shown in ~\cref{tab:cost}, \method{} demonstrates substantially faster inference, higher memory efficiency, and overall performance compared to other dynamic approaches.
Specifically, \method{} is highlighted by inference times that are almost 10 times faster than sample-wise dynamic interpolation methods such as TWIN-Merging~\cite{lu2024twin} and WEMoE~\cite{tang2024merging}. Our approach facilitates group batch processing for samples belonging to the same task ID. These results demonstrate the efficiency of our group batch processing strategy.

For the memory costs, \method{} significantly reduces memory requirements.
Existing dynamic merging methods face substantial memory challenges because they necessitates storing a complete task-specific model for every potential task. This results in a linear increase in memory cost with the number of merged tasks.
\method{}, however, combines with merging methods that leverage lightweight binary masks or compressed task experts, which store only compact task-specific updates rather than full task-specific models. This design substantially reduces memory cost, requiring memory comparable to static single-model merging methods, while achieving state-of-the-art performance that surpasses dynamic methods.
Furthermore, we provide a detailed analysis comparing our hard-routing strategy against soft-routing alternatives in terms of both end-to-end latency and accuracy in Appendix~\cref{app:efficiency}.

\begin{figure}[t]
\begin{minipage}[t]{0.46\textwidth}
  \centering
  \includegraphics[width=\linewidth]{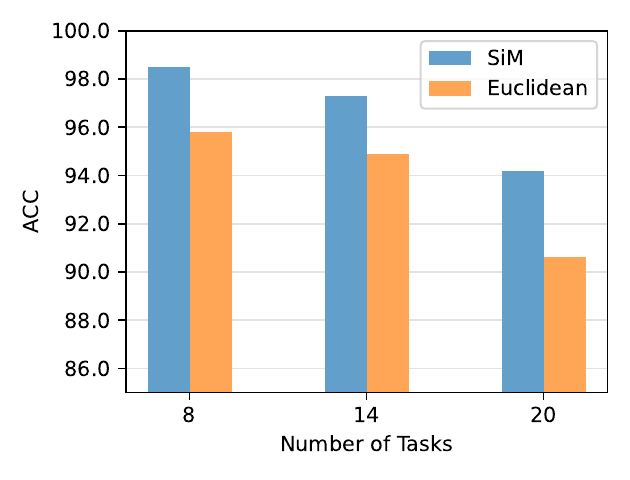}
  \caption{\textbf{Task classification performance of~\method{} vs Euclidean on the CLIP ViT-B/32 across 8, 14, and 20 tasks}}
  \label{fig:euclidean}
\end{minipage}\hfill
\begin{minipage}[t]{0.46\textwidth}
  \includegraphics[width=\linewidth]{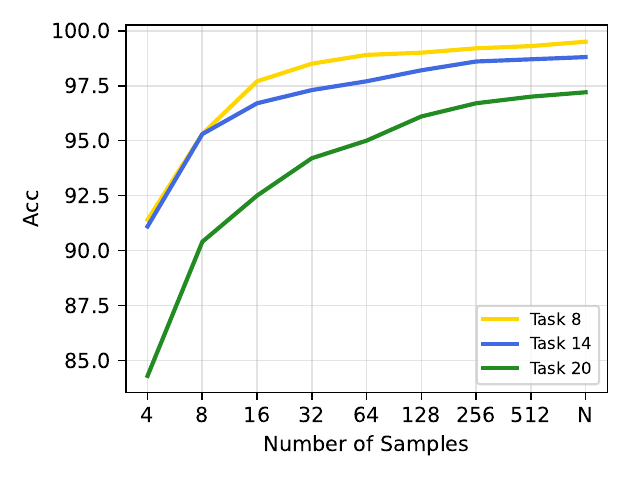}
  \caption{\textbf{Task classification accuracy using the CLIP ViT-B/32 across sample sizes}}
  \label{fig:classification}
\end{minipage}
\end{figure}

\subsection{Task classification}
\label{sec:task_cls}
\noindent\textbf{Projection residual vs Euclidean.}
As discussed in \cref{sec:theoretical analysis}, projection residuals enhance task separability by measuring alignment with respect to task-specific subspaces rather than relying solely on mean-based distances.
This observation is further supported by \cref{tab:ablation}, providing a comparison of training-free task classification performance across various distance-based metrics on the 8 vision task scenarios.
Overall, \method{} achieves superior performance across datasets, demonstrating that improved separability in residual space provides a more reliable signal for task classification. Furthermore, \method{} exhibits strong scalability, maintaining robust performance even as the number of tasks increases to 20, which leads to a widening performance gap compared to the Euclidean baseline, as illustrated in~\cref{fig:euclidean}.
Notably, on EuroSAT, where the diagonal entry in \cref{fig:ratio} (i.e., the residual ratio evaluated on the corresponding task subspace) is the smallest, \method{} achieves over a 10\% improvement compared to Euclidean and cosine baselines in \cref{tab:ablation}.
Moreover, we observe consistent gains on GTSRB and RESISC45, which also exhibit relatively small diagonal residual ratios, suggesting that smaller in-task residual ratios provide stronger separability in residual space.
Furthermore, \method{} outperforms the Mahalanobis distance, which accounts for covariance structure and non-parametric $k$-NN.
These results validate the effectiveness of \method{} as a training-free task classification approach and confirm that modeling task-specific subspaces yields a more discriminative and robust criterion for task classification.
Notably, as detailed in Appendix~\cref{app:routingerror}, even when routing errors occur, their effect on downstream performance is minimal. Replacing \method{} with oracle routing yields only a 0.3–0.6$\%$ performance gap, demonstrating that \method{} remains highly robust at the downstream prediction level.

\noindent\textbf{The number of samples.}
We have investigated the impact of sample size on the performance of \method{} when estimating task manifolds. 
\textcolor{black}{
As shown in \cref{fig:classification}, \method{} consistently achieves superior classification performance across diverse task scales, eliminating the need for any additional training.
}
Furthermore, \method{}, even with only 32 samples per task, yields strong performance similar to that achieved using the entire training dataset. Here, \(N \) represents the total number of training samples. This advantage is consistently demonstrating its robustness and broad applicability. 
A detailed task classification accuracy for each individual dataset across varying numbers of support samples is provided in Appendix~\cref{app:support-set}.
Additionally, as detailed in Appendix~\cref{app:block}, SiM can achieve near-optimal task classification using features from intermediate transformer blocks, offering a practical way to further reduce computational overhead without sacrificing performance.

\section{Conclusion}
In this work, we introduce \method , a training-free framework for dynamic model merging in task-agnostic scenarios.
By utilizing SVD-based manifold projections, our approach effectively identifies task IDs without any additional router training.
Furthermore, by integrating with subspace-based merging, \method{} eliminates the need to store full expert parameters, significantly reducing memory overhead.
Our experiments across vision and NLP benchmarks demonstrate that \method{} substantially improves performance and consistently narrows the gap to individual task experts in real-world settings where task information is unavailable.

\section*{Acknowledgments}
This work was supported by the Institute of Information and communications Technology Planning and evaluation (IITP) grant (No. RS-2025-25422680, No. RS-2020-II201373, Artificial Intelligence Graduate School Program (Hanyang University)) and the National Research Foundation of Korea (NRF) grant funded by the Korea government (MSIT) (No. RS-2025-24533064).
\newpage
%
%
\bibliographystyle{splncs04}
\bibliography{main}

\clearpage

\setcounter{section}{0}
\setcounter{figure}{0}
\setcounter{table}{0}
\setcounter{equation}{0}
\renewcommand{\thesection}{\Alph{section}}
\renewcommand{\thefigure}{\Alph{figure}}
\renewcommand{\thetable}{\Alph{table}}
\renewcommand{\theequation}{\Alph{equation}}

\renewcommand{\theHsection}{\Alph{section}}
\renewcommand{\theHfigure}{\Alph{figure}}
\renewcommand{\theHtable}{\Alph{table}}
\renewcommand{\theHequation}{\Alph{equation}}

\appendix
\section*{Appendix}
\section{Table of contents}
We provide the following items in this Appendix:
\begin{itemize}
    \item (\cref{app:implemantation}) Implementation details of \method
    \begin{itemize}
        \item (\cref{app:datausage}) Ephemeral data usage in the standard merging pipeline
        \item (\cref{app:code}) Pseudocode
    \end{itemize}
    \item (\cref{app:experimental}) Experimental details
    \begin{itemize}
        \item (\cref{app:train_detail}) Expert model preparation
        \item (\cref{app:baseline_detail}) Baseline details
        \item (\cref{app:inf_detail}) Inference details
    \end{itemize}
    \item (\cref{app:additonal}) Additional experiments on task classification
    \begin{itemize}
        \item (\cref{app:larger}) Scalability to larger task sets
        \item (\cref{app:similar}) Performance on fine-grained and similar tasks
        \item (\cref{app:block}) Performance across transformer blocks
    \end{itemize}
    \item (\cref{app:moreablation}) More ablation and analysis studies
    \begin{itemize}
        \item (\cref{app:ab_rank}) Varying rank $k$
        \item (\cref{app:ab_seed}) Varying seeds and sample sizes
        \item (\cref{app:support-set}) Support-set quantity and quality
        \item (\cref{app:efficiency}) Inference efficiency and routing variants
        \item (\cref{app:routingerror}) Effect of routing errors on downstream performance

    \end{itemize}
    \item (\cref{app:limitations}) Limitations
\end{itemize}
\section{Implementation details of \method}
\label{app:implemantation}

\subsection{Ephemeral data usage in the standard merging pipeline}
\label{app:datausage}
To clarify the role of \method{} in practical deployment, we situate it within the standard pipeline of expert construction and model merging, as illustrated in Table~\ref{tab:datausage}.
Within this pipeline, most model merging frameworks utilize task-specific data primarily during Stage 2 to fine-tune experts. Once these experts are prepared at Stage 3, the subsequent merging processes in Stages 4--5 are ideally intended to be data-independent.

However, as shown in Table~\ref{tab:datausage}, existing routing-based approaches~\cite{lu2024twin,tang2024merging,ye2025dynamic} typically require additional, persistent data in Stages 4--5 to train routers. Such requirements can be restrictive in practice, especially when the original training data are unavailable due to privacy, licensing, or storage constraints.
In contrast, \method{} shifts the data dependency to Stage 2 (Experts preparation). By utilizing data ephemerally to construct task manifolds from the pretrained backbone, \method{} ensures that the subsequent merging and inference stages (Stages 4--5) remain entirely training-free and data-free. This shift makes our approach more practical for real-world scenarios where data access is restricted after the experts are released.

\begin{table*}[t]
    \centering
    \caption{\textbf{Comparison of data usage in standard merging pipelines.} We illustrate the five operational stages of merging and contrast the data dependency of each method. 
    Stages 1--3 (access to pre-trained model, task expert preparation and release) are preliminary stages, which are often assumed to be completed before the merging processes proceed.
    While routing-based methods require additional, persistent data usage for router training in the merging process and inference (Stage 4--5), \method{} utilizes data ephemerally during expert preparation (Stage 2) to construct the manifold, ensuring a training-free workflow in all subsequent stages.}

  \resizebox{1\textwidth}{!}{
    \begin{tabular}{l|c|c|c|c|c}
    \toprule
         & \textbf{Stage 1} & \textbf{Stage 2} & \textbf{Stage 3} & \textbf{Stage 4} & \textbf{Stage 5}\\
        \midrule
        \textbf{Standard} & Pretrained model  & Experts preparation  & Experts release & Merging process & Inference using \\
         \textbf{merging pipelines} & given & (training of experts) & (e.g., Huggingface) & (combine experts) & merged model\\
         \midrule
         \textbf{Data usage for task expert preparation }  & \multirow{2}{*}{-}  & \multirow{2}{*}{\mycheckmarkrev}  & \multirow{2}{*}{-} & \multirow{2}{*}{-} & \multirow{2}{*}{-} \\
         \textbf{for all merging methods} &  &  & &  \\
         \midrule
         \rowcolor[HTML]{EFEFEF}
        \multicolumn{6}{l}{\textit{\textbf{(Additional data usage comparison)}}} \\
        \textbf{Routing-based} & \multirow{2}{*}{-}  & \multirow{2}{*}{-}  & \multirow{2}{*}{-} & \multicolumn{2}{c}{\mycheckmark (many)} \\
         \textbf{merging~\cite{lu2024twin, tang2024merging, ye2025dynamic}} & &  & &\multicolumn{2}{c} {Additional router training}\\
         \midrule
         \multirow{2}{*}{\textbf{\method}} & \multirow{2}{*}{-}  & \mycheckmarkrev (few)  & \multirow{2}{*}{-}  & \multirow{2}{*}{-}  & \multirow{2}{*}{-}  \\
          &   & Manifold construction  &  &  &  \\
          \bottomrule
    \end{tabular}
    }
\label{tab:datausage} 
\end{table*}

\subsection{Pseudocode}
\label{app:code}
We provide the formal procedure of \method{} in~\cref{alg:proposed_eff}. The pseudocode clearly reflects our design philosophy of shifting data dependency to the earlier preparation stage to achieve a seamless, training-free merging process.

\setlength{\intextsep}{2pt}
\setlength{\columnsep}{10pt}

\begin{algorithm}[t]\small
    \caption{\method}\label{alg:proposed_eff}
    \begin{algorithmic}[1]
    \Require 
    Pretrained weight $\vtheta_0$,
    compressed task vectors \( \{\mathbf{\Delta}_t\}_{t=1}^T \),
    training samples $\{{\vx}_i^{(t)}\}_{i=1}^N$ per task \(t\),
    rank $k$ (for top-\(k\) singular vectors)
    \Ensure 
    Final model outputs $\boldsymbol{Y}$ for all test inputs

    \Statex
    \State \textbf{Step 1: Task manifold pre-computation.}
    \Comment{{\color{blue} Before merging}}
    \For{$t = 1$ to $T$}
            \State Extract features:
            \State $\quad
                f_\text{pre}({\vx}_i^{(t)}) = f({\vx}_i^{(t)}; \boldsymbol{\theta}_\text{0}), \quad i=1,\dots,N
            $
            \State Compute \( {\bm\mu}_t \) and \( {\mathbf{V}}_t \) as in Eq. (\ref{eq:mean}) and Eq. (\ref{eq:singular})
            \State Retain the top-$k$ singular vectors to form \( {\mathbf{V}}_{t,k} \)
        \EndFor
    \Statex

    \State \textbf{Step 2: Group batch inference.}
    \Comment{{\color{blue}Test time (Inference)}}
    \State Initialize output $\boldsymbol{Y} \leftarrow \{\}$
    \For{each test batch $\acute{\boldsymbol{X}}$}
        \State Extract pretrained features 
        $\mathbf{Z} = f_{\mathrm{pre}}(\acute{\boldsymbol{X}}) \in \mathbb{R}^{B \times d}$
        \For{$t = 1$ to $T$}
            \State Compute \( R_t({\boldsymbol{\mathbf{Z}}})\) as in Eq. (\ref{eq:residual})
        \EndFor
        \State \( \hat{t} = \argmin_{t\in\{1,\ldots,T\}} R_t({\boldsymbol{Z}}) \)
        \State Group \( \acute{\bm{X}} \) by predicted task IDs $\hat{t}$
        \For {each predicted task \( \hat{t}\)}
            \State \( \vtheta^\star_\text{merge} \gets \vtheta_0 + \mathbf{\Delta}_{\hat{t}}\)
            \State Forward pass for the \( \hat{t} \)-th input group \( \boldsymbol{X}_{\hat{t}}' \) using \( \boldsymbol{\theta}^\star_\text{merge} \): \( f(\boldsymbol{\acute{X}}_{\hat{t}}; \vtheta^\star_\text{merge}) \)
            \State \( \boldsymbol{Y} \gets \boldsymbol{Y} \cup f(\boldsymbol{\acute{X}}_{\hat{t}}; \vtheta^\star_\text{merge}) \)
        \EndFor
    \EndFor   
    \end{algorithmic}
\end{algorithm}

\section{Experimental details}
\label{app:experimental}

\subsection{Expert model preparation}
\label{app:train_detail}
\noindent\textbf{Expert model acquisition.}
We obtain the initial expert models following the aforementioned TALL-Mask~\cite{wang2024localizing} training settings. Building upon these experts, we have subsequently applied various methods—including EMR-Merging~\cite{huang2024emr}, TALL-Mask~\cite{wang2024localizing}, and TSV-C~\cite{gargiulo2024task}—to generate compressed task vectors or task-specific binary masks for our multi-task integration experiments.

\noindent\textbf{Fine-tuning details for computer vision tasks (21--50 tasks).}
For the experiments involving 21 to 50 tasks, we have independently fine-tuned a separate expert model for each task to obtain the initial expert set. Following the training methodology outlined by TALL-Mask~\cite{wang2024localizing},
we have employed the same pre-trained CLIP ViT~\cite{dosovitskiy2021an, radford2021llearning} checkpoint sourced from the \texttt{open\_clip} repository~\cite{ilharco2023editing}, fine-tuning for 2,000 iterations.
The dataset has been split into training and validation sets, comprising 90\% and 10\% of the data, respectively.
The training configuration included a batch size of 128, a learning rate of $10^{-5}$ (managed by a cosine annealing schedule with 200 warm-up steps), the AdamW optimizer, and cross-entropy as the loss function.
Furthermore, in accordance with Task Arithmetic~\cite{ilharco2023editing} and TALL-Mask~\cite{wang2024localizing}, the classification layer weights have remained fixed throughout fine-tuning.

\subsection{Baseline details}
\label{app:baseline_detail}
We now describe the baselines utilized for our main comparison experiment:
\begin{itemize}
    \item \textbf{Fine-tuned}: This baseline represents a fine-tuned task-specific model for each task.
    These models are inherently single-task (i.e., they cannot execute multiple tasks concurrently). 
    \item \textbf{Task Arithmetic}~\cite{ilharco2023editing}: This baseline first computes \textit{task vectors} \( \boldsymbol{\tau}_t \) for each task \( t \) by determining the difference between its fine-tuned parameters \( \vtheta_t \) and the pre-trained parameters \( \vtheta_0 \) (i.e., \( \boldsymbol{\tau}_t = \vtheta_t - \vtheta_0 \)).
    A merged model \( \vtheta_\text{merge} \) is then formed by adding a linear combination of these task vectors to the pre-trained parameters.
    This combination is expressed as \( \vtheta_\text{merge} = \vtheta_0 + \alpha \sum_{t=1}^T \boldsymbol{\tau}_t  \), where each task vector \( \boldsymbol{\tau}_t \) is scaled by a hyper-parameter \( \alpha \).
    The hyper-parameter \( \alpha \) is tuned from the set \( \{0.0, 0.1, ..., 0.9, 1.0\} \) by optimizing for the average performance across all tasks on their respective validation sets.
    \item \textbf{TIES-Merging}~\cite{yadav2023ties}: This baseline performs weight interpolation through three stages: \textit{Trim}, \textit{Elect Sign}, and \textit{Merge}.
    During the Trim stage, all task vector values are set to zero except for the top 20\% based on magnitude.
    The Merge stage then proceeds identically to Task Arithmetic, with hyper-parameter tuning conducted as described previously.
    \item \textbf{TSV-M}~\cite{gargiulo2024task}: This baseline leverages the low-rank structure of per-layer task matrices $\mathbf{\Delta}_i$ by retaining only the top-$k$ (e.g., 1/$T$) singular components via SVD. To mitigate task interference, it applies a whitening transformation by solving the Orthogonal Procrustes problem, yielding decorrelated orthogonal matrices $U_{\perp}$, $V_{\perp}$. The merged update is then reconstructed as $\hat{M}=U_{\perp} \Sigma V_{\perp}$ and added to the pre-trained model $\vtheta_0$ with a scaling factor $\alpha$, following the same hyper-parameter tuning as previous baselines.
    \item \textbf{Iso-C}~\cite{iso2024daniel}: This baseline performs model merging by flattening the singular value spectrum of the aggregated task matrix $\mathbf{\Delta}_{\text{TA}}$. After decomposing $\mathbf{\Delta}_{\text{TA}}$ via SVD into $U\Sigma V^T$, it replaces all singular values with their average $\bar\sigma=\frac{1}{r}\Sigma\sigma_i$, to achieve an isotropic distribution. The merged update is then reconstructed as $\mathbf{\Delta}_{\text{Iso-C}}=\bar\sigma U V^T$ and added to the pre-trained weights $\vtheta_0$  with a scaling factor $\alpha$.
    \item \textbf{TWIN-Merging}~\cite{lu2024twin}: This baseline initially trains a router \( \mathcal{R}(\cdot; \phi) \), parameterized by \( \phi \), to enable effective task classification.
    Subsequently, a \textit{shared expert} \( f(\cdot; \vtheta_s) \) parameterized by \( \vtheta_s \) is extracted by Task Arithmetic with fixed hyper-parameter (\( \alpha = 0.29 \)).
    Following this, \textit{exclusive knowledge vectors} \( \bm{v}_t \) are extracted for each task, employing either Singular Value Decomposition (SVD) or the Trim procedure from TIES-Merging.
    In this comparison, Trim demonstrates the best performance.
    During test time, the router determines task-specific weights \( w_t = \operatorname{softmax}(\mathcal{R}(\text{Emb}(\vx); \phi)) \) for each input \( \vx \), where \( \text{Emb}(\vx) \) denotes the embeddings of the penultimate layer from the shared expert.
    Weight interpolation is then performed by adding to the shared expert a weighted sum of these exclusive knowledge vectors, where the weights are the routed task-specific weights: \( \vtheta_\text{merge} = \vtheta_s + \sum_{t=1}^T w_t \bm{v}_t\).
    \item \textbf{DaWin}~\cite{oh2025dawin}: This baseline performs weight interpolation by assigning a weight to each task for every input.
    This weight \( \lambda_t(\vx) \) for task \( t \) is derived from an entropy calculation that considers the task-specific model for that input \( x \).
    Weight interpolation is then performed according to the following equation: \( \vtheta_\text{merge} = \vtheta_0 + \alpha \sum_{t=1}^{T} \lambda_t(\vx) \bm\tau_t \), where \( \alpha \) is fixed at $0.3$.
    Additionally, to reduce runtime overhead during inference, a Beta mixture model (BMM) can also be employed.
    The default number of mixture components (\( K \)) for BMM is 3.
    \item \textbf{WEMoE}~\cite{tang2024merging}: This baseline upscales the MLP layers of Transformer blocks into a Weight-Ensembling Mixture of Experts module. It separates shared knowledge ($\vtheta_0^{\text{MLP}}$) from task-specific knowledge (encoded as a dictionary of task vectors, $\mathbf{D}_{\tau}$). For each input, a router $r$ generates routing weights $w$, and the dynamic MLP parameters are reconstructed as $\vtheta^{\text{MLP}} = \vtheta_0^{\text{MLP}}+\mathbf{D}_{\tau}w$, allowing the model to adaptively integrate knowledge based on the input instance. To optimize the router without downstream labels, the method employs test-time adaptation via entropy minimization on unlabeled data.
    \item \textbf{MoW-Merging}~\cite{ye2025dynamic}: This baseline performs sample-wise weight fusion using a lightweight gating network. It generates input-specific merging coefficients $p_i$ to adaptively combine model weights as $\mathbf{W}^l_M=\sum p_iW_i^l$. To balance computational efficiency and performance, the method employs a weight similarity metric to identify layers with severe interference; dynamic merging is applied only to these selected layers, while layers with high similarity are merged using static methods.
    \item \textbf{EMR-Merging}~\cite{huang2024emr}: This baseline first obtains an aggregated \textit{unified task vector} \( \bm\tau_\text{uni} \) by leveraging the sign and absolute value of each individual task vector.
    Subsequently, it computes a task-specific mask \( M_t \) and rescaler \( \lambda_t \) for each task.
    The task-specific mask and rescaler are adjusted according to the particular evaluation task.
    Finally, weight interpolation is achieved by element-wise multiplying the unified task vector with the respective task-specific masks and rescalers: \( \vtheta_\text{merge} = \vtheta_0 + \lambda_t \cdot M_t \odot \bm\tau_\text{uni} \).
    Notably, this baseline does not involve any hyper-parameter tuning.
    \item \textbf{TALL Mask}~\cite{wang2024localizing}: This baseline initially employs a task vector of multi-task model \( \boldsymbol{\tau}_\text{MTL} \) to generate masks that identify parameters crucial for each task.
    Typically, the multi-task model is created using Task Arithmetic (TA) or TIES-Merging (TIES).
    When using TA, the scaling factor \( \alpha \) is tuned within \( \{0.0, 0.1, \ldots, 1.0\} \) and selected according to the average validation performance over all tasks.
    Subsequently, a \textit{task localization mask} \( \boldsymbol{m}_t \) is created to construct \(\boldsymbol{\hat\theta}_t\) such that:
    \(\boldsymbol{\hat\theta}_t\ = \boldsymbol{\theta}_0 + \boldsymbol{m}_t \circ \boldsymbol{\tau}_{\text{MTL}}\).
    The mask is obtained by minimizing the \(\ell_1\) distance between the reconstructed \(\hat\theta_t\) and fine-tuned parameters \(\theta_t\).
    The generation of these masks is governed by a hyper-parameter \( \lambda_t \) for task \( t \), which is tuned over the range \( \{0.2, 0.3, 0.4, 0.5, 0.6\} \) based on validation set performance.
    \item \textbf{TSV-C}~\cite{gargiulo2024task}: This baseline performs task vector compression by leveraging the inherently low-rank structure of per-layer task matrices. For each task $i$, it applies Singular Value Decomposition (SVD) to the task matrix $\mathbf{\Delta}_i$ and retains a reduced number $k$ (e.g., $k\approx \text{rank}/T$) of the most significant singular components.  The compressed update is reconstructed as $\mathbf{\hat\Delta}_i=\sum_{j=1}^k\sigma_j^iu_j^iv_j^i$, allowing the model to achieve a flexible trade-off between storage efficiency and task-specific fidelity.
    
\end{itemize}

\subsection{Inference details}
\label{app:inf_detail}
For all inference procedures, including performance evaluation and computational cost measurement, experiments involving 14 tasks or fewer have been conducted using NVIDIA GeForce RTX 3090 GPUs, while those with 20 or more tasks have utilized NVIDIA H100 80GB HBM3 GPUs.



\section{Additional experiments on task classification }
\label{app:additonal}

\subsection{Scalability to larger task sets}
\label{app:larger}
To assess scalability to larger and more heterogeneous task collections, we additionally evaluate task classification on 30 and 50 vision tasks using a CLIP ViT-B/32 backbone.
Specifically, the 30-task set extends the initial 20 with:
Vegetables~\cite{ahmed2021dcnn}, 
Kvasir-v2~\cite{pogorelov2017kvasir}, 
Intel Images~\cite{bansal2019intel}, 
Weather~\cite{xiao2021classification}, 
Cats and dogs~\cite{cukierski2013dogs}, 
MangoLeafBD~\cite{ahmed2023mangoleafbd}, 
Beans~\cite{beansdata}, 
Landscape Recognition~\cite{Landscape}, 
Garbage Classification~\cite{cchang_2018}, 
and Fruits-360~\cite{muresan2018fruit}.
The 50-task set further adds 20 more datasets, including: 
PlantVillage~\cite{mohanty2016plantdisease},
OCT2017~\cite{kermany2018identifying},
ASL Alphabet~\cite{grassknoted_asl_alphabet},
AID~\cite{xia2017aid},
CUB200-2011~\cite{wah2011cub},
ObjectCategories-256~\cite{griffin2007caltech256},
Animals-10~\cite{corrado_animals10},
EyePACS~\cite{cuadros2009eyepacs},
StanfordDogs~\cite{khosla2011fgvc},
MIT Indoor-67~\cite{quattoni2009indoor},
Imagenette2-320~\cite{howard2019imagenette},
Caltech-101~\cite{fei2007caltech101},
RAF-DB~\cite{li2017reliable},
FGVC-Aircraft~\cite{maji2013aircraft},
DeepWeeds~\cite{olsen2019deepweeds},
Imagewoof~\cite{howard2019imagenette},
NABirds~\cite{vanhorn2015nabirds},
Leafsnap~\cite{kumar2012leafsnap},
AFHQ~\cite{choi2020starganv2},
ArtBench10~\cite{liao2022artbench}.

\begin{figure}[t]
    \centering
    \includegraphics[width=0.8\linewidth]{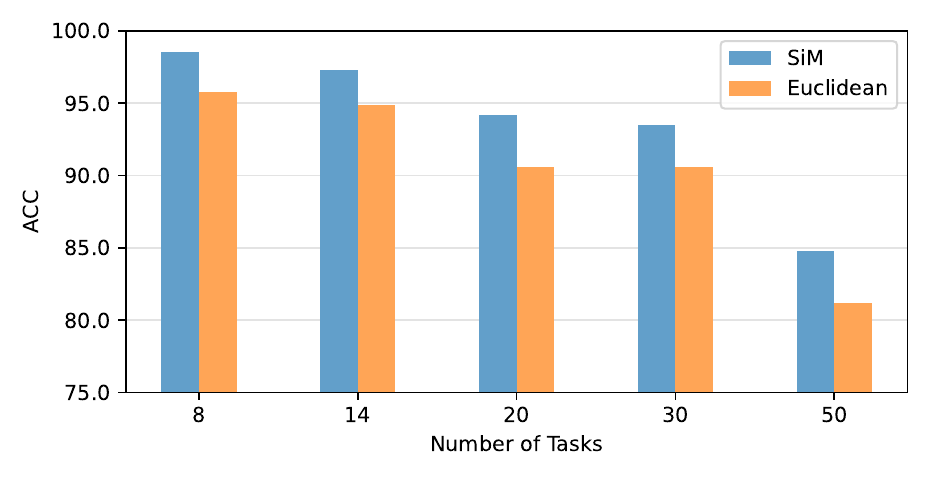}
  \caption{\textbf{Task classification performance of~\method{} vs Euclidean on the CLIP ViT-B/32 for 30 and 50 tasks}}
  \label{fig:3050classification}
\end{figure}

As shown in~\cref{fig:3050classification}, while the classification problem is inherently more challenging as the number of tasks increases, \method{} maintains a consistently superior and stable margin over the Euclidean distance baseline. This stability across growing task counts suggests that \method{} effectively captures discriminative task-specific features that distance-based metrics fail to resolve in high-dimensional task spaces.
This robust classification capability directly contributes to superior multi-task performance. As shown in~\cref{tab:3050}, conventional merging methods experience a significant performance degradation in these large-scale settings. However, when \method{} is integrated with existing subspace-based merging methods, it achieves the highest performance for both 30 and 50 task scenarios. These results collectively demonstrate that \method{} is a highly robust and scalable solution, capable of maintaining high-fidelity performance as the complexity and variety of tasks grow.

\begin{table*}[t!]
\centering
\caption{
\textbf{Multi-task performance for 30 and 50 computer vision tasks.} We report experimental results for ViT-B/32 on 30 and 50 vision tasks. Bold values represent the best performance among all methods, excluding the individual task-specific baselines.}

\label{tab:3050}
\setlength\tabcolsep{4.0pt}
\resizebox{0.75\textwidth}{!}{
\begin{tabular}{l|cc}
\toprule
\textbf{Method} 
& \textbf{30 tasks} 
& \textbf{50 tasks} \\
\midrule
Fine-tuned & 93.1 & 90.8 \\
\midrule
\rowcolor[HTML]{EFEFEF}
\multicolumn{3}{l}{\textbf{\textit{(Static model merging)}}} \\
Task Arithmetic~\cite{ilharco2023editing} & 58.0 & 46.8\\
TIES-Merging~\cite{yadav2023ties}    & 59.6 & 47.5\\
TSV-M~\cite{gargiulo2024task}    & 77.3 & 66.6\\
Iso-C~\cite{iso2024daniel}    & 72.8 & 49.1\\
\midrule
\rowcolor[HTML]{EFEFEF}
\multicolumn{3}{l}{\textbf{\textit{(Dynamic model merging)}}} \\
TWIN-Merging~\cite{lu2024twin} & 60.1 & 36.4\\
DaWin~\cite{oh2025dawin}  & 40.3 &  33.1 \\
WEMoE~\cite{tang2024merging} & 67.1 & 51.6 \\
MoW-Merging~\cite{ye2025dynamic}   & 56.4 & 30.1 \\
\midrule
\rowcolor[HTML]{FFF2CC}
\textbf{EMR~\cite{huang2024emr} + \method{} (Ours) }                 & 89.0 & 76.7 \\
\rowcolor[HTML]{FFF2CC}
\textbf{TM-TA~\cite{wang2024localizing} + \method{} (Ours)}                        & 87.1 & 73.1\\
\rowcolor[HTML]{FFF2CC}
\textbf{TM-TIES~\cite{wang2024localizing} + \method{} (Ours) }                     & 85.6 & 72.2 \\
\rowcolor[HTML]{FFF2CC}
\textbf{TSV-C~\cite{gargiulo2024task} + \method{} (Ours) }                     & \textbf{90.0} & \textbf{80.2} \\
\bottomrule
\end{tabular}
}
\end{table*}

\begin{figure}[t]
\centering
\begin{minipage}[t]{0.48\textwidth}
  \centering
  \includegraphics[width=0.9\linewidth]{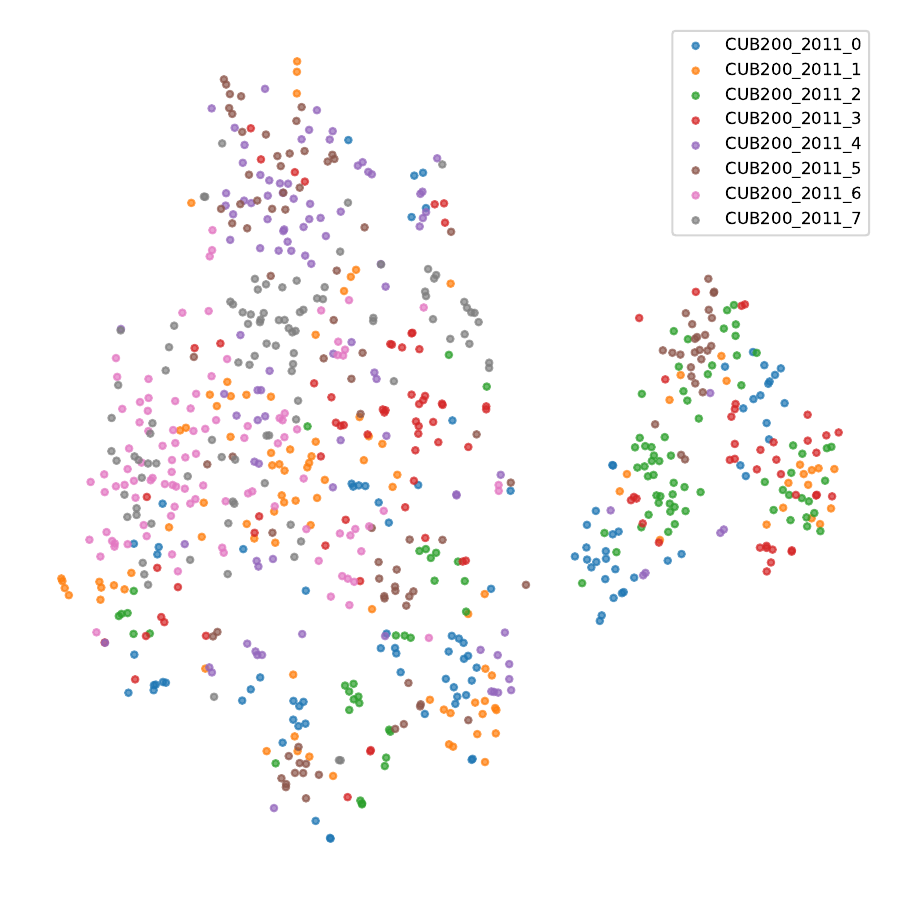}
  \caption{\textbf{Task-specific feature distributions from the CLIP ViT-B/32 across 8 fine-grained bird classification tasks (CUB-200 dataset~\cite{wah2011cub})}}
  \label{fig:bird}
\end{minipage}\hfill
\begin{minipage}[t]{0.48\textwidth}
  \centering
  \includegraphics[width=\linewidth]{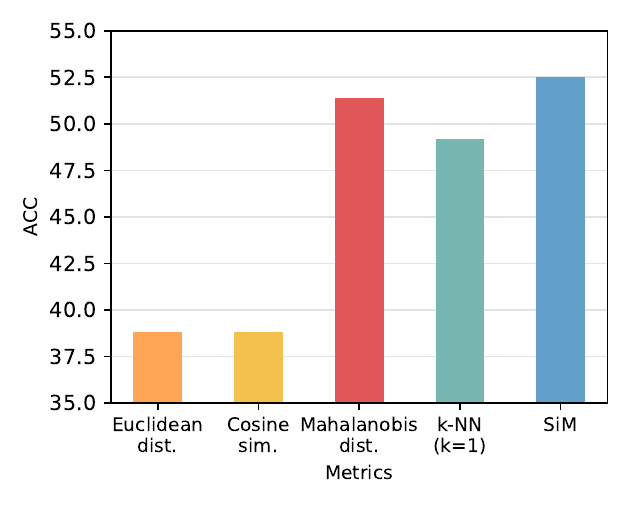}
  \caption{\textbf{Task classification performance comparison across different distance-based metrics for 8 fine-grained bird classification tasks using 32 support samples ($N=32$)}}
  \label{fig:bird_tc}
\end{minipage}
\end{figure}

\begin{table*}[t!]
\centering
\caption{
\textbf{Multi-task performance for 8 fine-grained bird classification tasks (CUB-200 dataset~\cite{wah2011cub}).} We report experimental results for ViT-B/32 on 8 fine-grained bird classification tasks from the CUB-200 dataset. Bold values represent the best performance among all methods, excluding the individual task-specific baselines.}

\label{tab:bird}
\setlength\tabcolsep{4.0pt}
\resizebox{0.65\textwidth}{!}{
\begin{tabular}{l|c}
\toprule
\textbf{Method} 
& \textbf{Average accuracy}  \\
\midrule
Fine-tuned & 78.1 \\
\midrule
\rowcolor[HTML]{EFEFEF}
\multicolumn{2}{l}{\textbf{\textit{(Static model merging)}}} \\
Task Arithmetic~\cite{ilharco2023editing} & 62.1\\
TIES-Merging~\cite{yadav2023ties}    & 61.4\\
TSV-M~\cite{gargiulo2024task}    & 65.7\\
Iso-C~\cite{iso2024daniel}    & 61.8\\
\midrule
\rowcolor[HTML]{EFEFEF}
\multicolumn{2}{l}{\textbf{\textit{(Dynamic model merging)}}} \\
TWIN-Merging~\cite{lu2024twin} & 64.6\\
DaWin~\cite{oh2025dawin}  & 61.0\\
WEMoE~\cite{tang2024merging} & 57.0 \\
MoW-Merging~\cite{ye2025dynamic}   & 58.9 \\
\midrule
\rowcolor[HTML]{FFF2CC}
\textbf{EMR~\cite{huang2024emr} + \method{} (Ours) }                 & 65.5 \\
\rowcolor[HTML]{FFF2CC}
\textbf{TM-TA~\cite{wang2024localizing} + \method{} (Ours)}                        & \textbf{67.1}\\
\rowcolor[HTML]{FFF2CC}
\textbf{TM-TIES~\cite{wang2024localizing} + \method{} (Ours) }                     & 66.4 \\
\rowcolor[HTML]{FFF2CC}
\textbf{TSV-C~\cite{gargiulo2024task} + \method{} (Ours) }                     & 65.5 \\
\bottomrule
\end{tabular}
}
\end{table*}

\begin{figure}[t]
\centering
\begin{minipage}[t]{0.48\textwidth}
  \centering
  \includegraphics[width=0.9\linewidth]{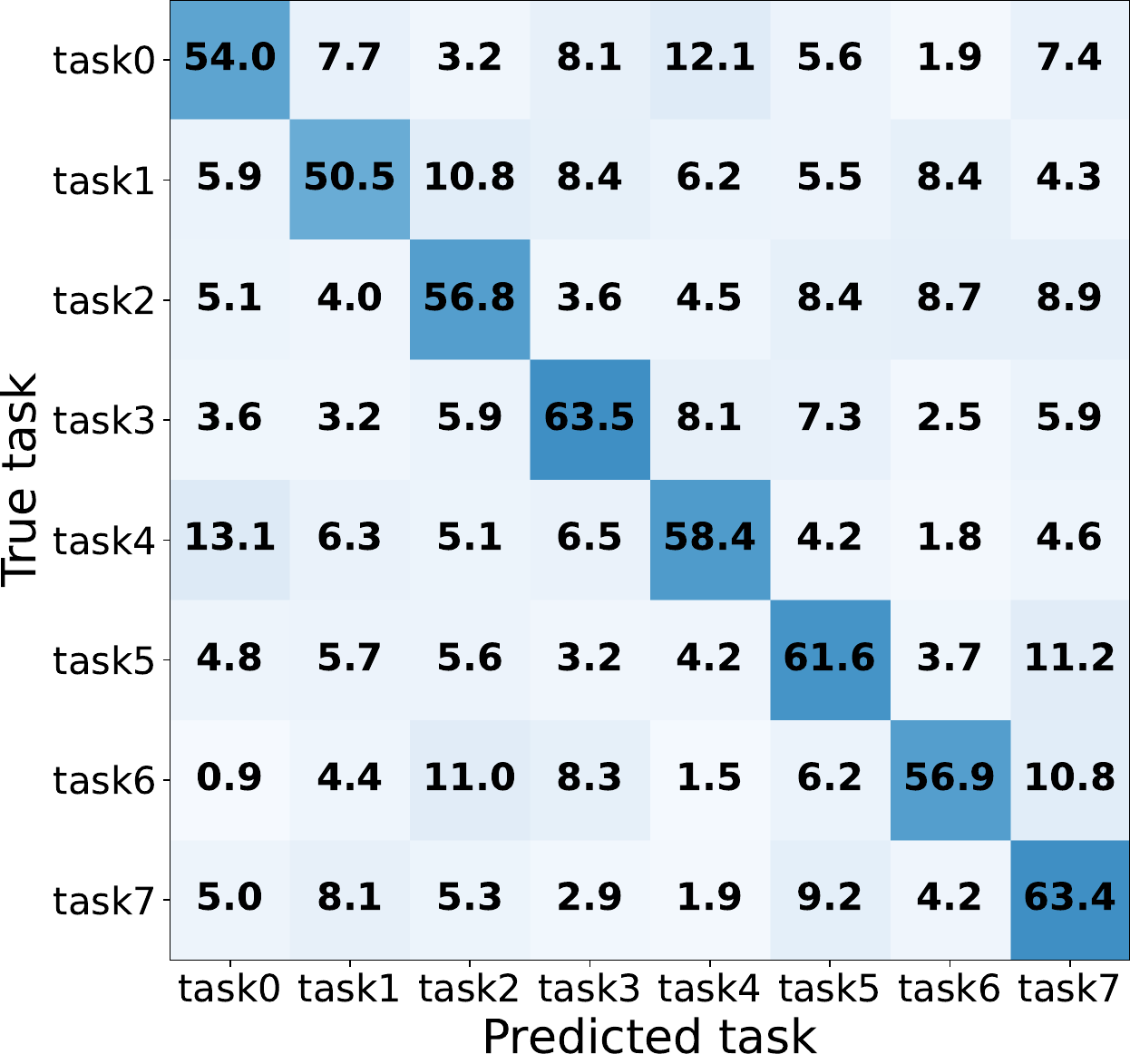}
  \caption{\textbf{Task classification confusion matrix from the CLIP ViT-B/32 across 8 fine-grained SUN-397~\cite{xiao2016sun}-based similar tasks}}
  \label{fig:sun_cf}
\end{minipage}\hfill
\begin{minipage}[t]{0.48\textwidth}
  \centering
  \includegraphics[width=\linewidth]{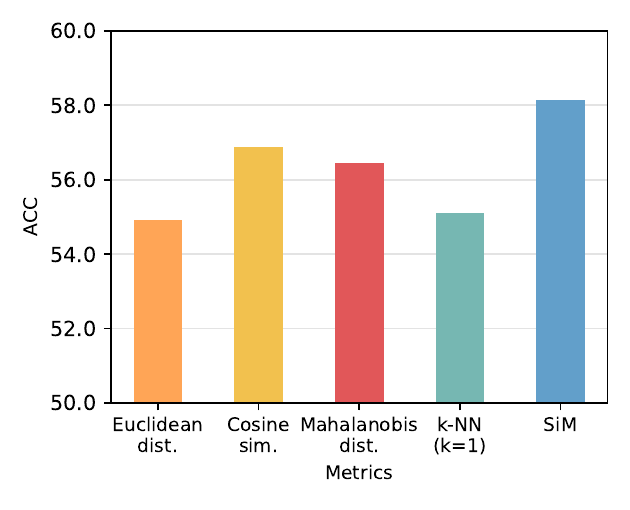}
  \caption{\textbf{Task classification performance comparison across different distance-based metrics for 8 fine-grained SUN-397~\cite{xiao2016sun}-based similar tasks using 32 support samples ($N=32$)}}
  \label{fig:sun_tc}
\end{minipage}
\end{figure}

\begin{table*}[t!]
\centering
\caption{
\textbf{Multi-task performance for 8 fine-grained SUN-397~\cite{xiao2016sun}-based similar tasks.} We report experimental results for ViT-B/32. Bold values represent the best performance among all methods, excluding the individual task-specific baselines.}

\label{tab:sun}
\setlength\tabcolsep{4.0pt}
\resizebox{0.65\textwidth}{!}{
\begin{tabular}{l|c}
\toprule
\textbf{Method} 
& \textbf{Average accuracy}  \\
\midrule
Fine-tuned & 95.0 \\
\midrule
\rowcolor[HTML]{EFEFEF}
\multicolumn{2}{l}{\textbf{\textit{(Static model merging)}}} \\
Task Arithmetic~\cite{ilharco2023editing} & 90.6 \\
TIES-Merging~\cite{yadav2023ties}    & 90.8\\
TSV-M~\cite{gargiulo2024task}    & 92.7\\
Iso-C~\cite{iso2024daniel}    & 92.5\\
\midrule
\rowcolor[HTML]{EFEFEF}
\multicolumn{2}{l}{\textbf{\textit{(Dynamic model merging)}}} \\
TWIN-Merging~\cite{lu2024twin} & 92.0\\
DaWin~\cite{oh2025dawin}  & 90.7\\
WEMoE~\cite{tang2024merging} & 90.1 \\
MoW-Merging~\cite{ye2025dynamic}   & 92.2 \\
\midrule
\rowcolor[HTML]{FFF2CC}
\textbf{TM-TIES~\cite{wang2024localizing} + \method{} (Ours) }                     & \textbf{93.1} \\
\bottomrule
\end{tabular}
}
\end{table*}

\subsection{Performance on fine-grained and similar tasks}
\label{app:similar}
To rigorously assess the robustness of \method{} when faced with highly similar and fine-grained domains, we have conducted experiments using the CUB-200 birds dataset~\cite{wah2011cub}. Specifically, we have partitioned the dataset into eight disjoint subsets of classes, treating each as an individual task, and fine-tuned separate experts for each.

As visualized in~\cref{fig:bird}, these tasks exhibit substantial feature overlap, posing a significantly greater challenge for task identification compared to general-purpose datasets. As shown in \cref{fig:bird_tc}, traditional distance-based metrics, such as Euclidean distance and Cosine similarity, struggle in this regime to resolve the subtle boundaries between tasks.
While Mahalanobis distance and $k$-NN provide significant improvements, \method{} achieves the highest overall accuracy. This suggests that by effectively capturing subspace structures, \method{} gains a further advantage in resolving fine-grained differences compared to traditional distance-based baselines.
Furthermore, as shown in \cref{tab:bird}, integrating \method{} with existing subspace-based merging frameworks yields the highest multi-task performance among all evaluated methods. These results underscore the robustness of our approach, proving that \method{} remains effective and reliable even in ambiguous scenarios where task-specific features are heavily entangled.

We further evaluate \method{} under a SUN397-based similar-task setting.
Specifically, we construct eight tasks from SUN397, where each task consists of 20 fine-grained and semantically similar scene classes. 
As shown in~\cref{fig:sun_cf}, the confusion matrix exhibits non-negligible off-diagonal entries, indicating that several samples are routed to semantically related but incorrect tasks. 
Nevertheless, the diagonal entries remain dominant, showing that \method{} still captures discriminative task-specific structures even under ambiguous task boundaries. 
The task classification comparison in~\cref{fig:sun_tc} further confirms this trend: \method{} outperforms conventional distance-based metrics. 
Moreover, as reported in~\cref{tab:sun}, integrating \method{} with TM-TIES achieves the best multi-task performance among all evaluated merging methods.
These results further demonstrate that \method{} remains robust not only on fine-grained object categories, but also on highly similar scene classification tasks.

\begin{figure}[t]
\centering
\begin{minipage}[t]{0.32\textwidth}
  \centering
  \includegraphics[width=\linewidth]{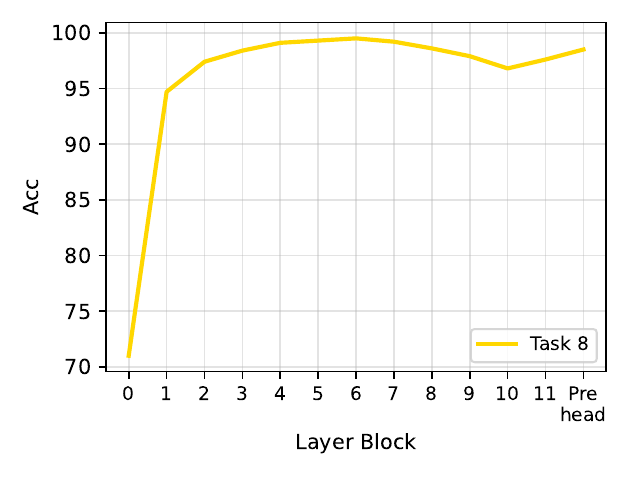}
  \caption{\textbf{Task classification performance across different layer block features from the CLIP ViT-B/32 for 8 vision tasks}}
  \label{fig:block8}
\end{minipage}\hfill
\begin{minipage}[t]{0.32\textwidth}
  \centering
  \includegraphics[width=\linewidth]{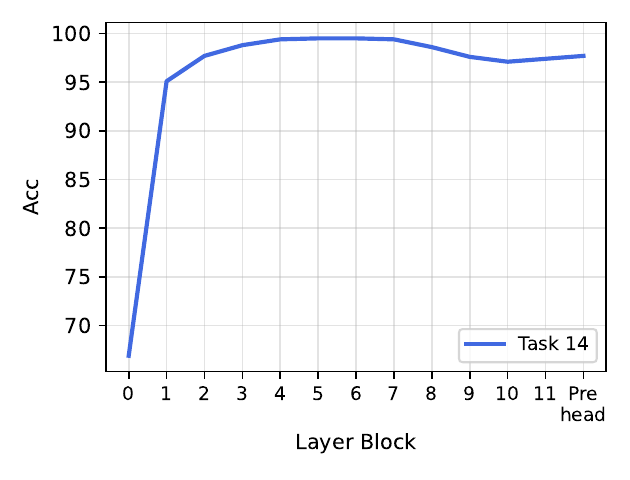}
  \caption{\textbf{Task classification performance across different layer block features from the CLIP ViT-B/32 for 14 vision tasks}}
  \label{fig:block14}
\end{minipage}\hfill
  \begin{minipage}[t]{0.32\textwidth}
  \centering
  \includegraphics[width=\linewidth]{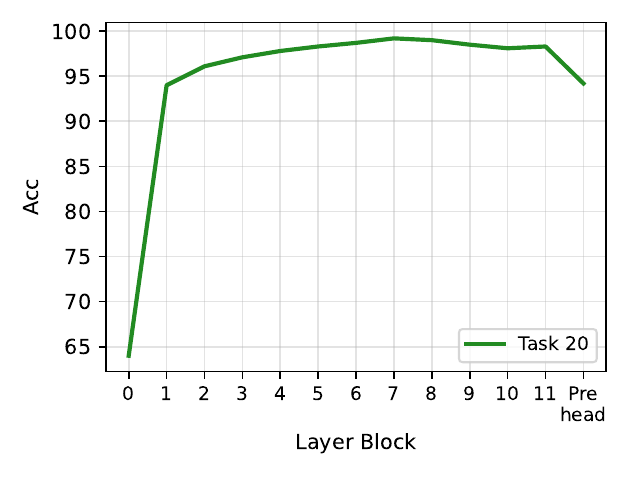}
  \caption{\textbf{Task classification performance across different layer block features from the CLIP ViT-B/32 for 20 vision tasks}}
  \label{fig:block20}
\end{minipage}
\end{figure}

\subsection{Performance across transformer blocks}
\label{app:block}
We further investigate the impact of the transformer block choice on task classification performance by extracting features at the entry of each block.
As illustrated in~\cref{fig:block8,fig:block14,fig:block20}, we observe a distinct trend where the classification accuracy rapidly increases within the initial layers and reaches its peak at the intermediate blocks—specifically between Block 5 and Block 7—across all task settings (8, 14, and 20 tasks).
Notably, these intermediate features consistently outperform those from the final layers, suggesting that mid-level representations in the CLIP ViT backbone capture more discriminative and generalized semantic information for task identification.

This observation provides significant practical advantages.
Since near-optimal performance is attained well before the final layer, task manifold construction does not strictly require a full forward pass through all 12 transformer blocks.
By leveraging early or intermediate features, we can construct task-specific subspaces more efficiently, reducing computational overhead while maintaining high classification accuracy. 
These results confirm that \method{} can be deployed flexibly to balance performance and computational efficiency.

\section{More ablation and analysis studies}
\label{app:moreablation}
\begin{table}[t!]
\centering
\caption{
\textbf{Ablation study on the impact of the rank $k$ on the task classification performance of \method{} using CLIP ViT-B/32 across 8, 14, and 20 tasks}. All results are obtained with 32 support samples per task ($N=32$).}
\label{tab:k_ablation}
\setlength\tabcolsep{4.5pt}
\resizebox{0.8\textwidth}{!}{
\begin{tabular}{c|cccccc}
\toprule
& $k=1$
& $k=2$
& $k=4$
& $k=8$ 
& $k=16$ 
& $k=32$  \\
\midrule
\textbf{8 tasks} & 98.3 & 98.4 & 98.6 & 98.6 & 98.6 & 98.5 \\
\textbf{14 tasks} & 96.8 & 97.5 & 97.6 & 97.6 & 97.7 & 97.7 \\
\textbf{20 tasks} & 92.8 & 93.2 & 93.7 & 94.1 & 94.3 & 94.2 \\
\bottomrule
\end{tabular}
}
\end{table}
\begin{table}[t!]
\centering
\caption{
\textbf{Ablation study on the impact of the sample size (\( N \)) and random seed initialization on the task classification performance of \method~using a CLIP ViT-B/32.}
Accuracy is reported across eight vision datasets.
The table shows results for \( N = \{1, 2, 4, 8, 16, 32\} \), where each entry represents the mean and its standard deviation (Mean $\pm$ std) calculated across five different random seeds (0–4) for each \( N \).}
\label{tab:seeds}
\setlength\tabcolsep{3.0pt}
\resizebox{\textwidth}{!}{
\begin{tabular}{c|cccccccc|c}
\toprule
\textbf{Method} & Cars & DTD & EuroSAT & GTSRB & MNIST & RESISC45 & SUN397 & SVHN & \textbf{Avg.} \\
\midrule
\method{} & 93.8  & 55.9  & 81.5  & 50.7  & 100  & 44.5  & 26.6 & 79.1 & 66.5  \\
 ($N=1$) & $\pm$ 3.8 & $\pm$ 20.2 & $\pm$ 12.2 &  $\pm$ 22.6 & $\pm$ 0.0 &  $\pm$ 29.1 & $\pm$ 4.9 & $\pm$ 18.2 &  $\pm$ 9.5 \\
\midrule
\method{} &99.6 & 83.1 & 88.4 & 79.6 & 100 & 58.1 & 56.2 & 88.8 & 81.7 \\
 ($N=2$) &$\pm$ 0.2 &$\pm$ 10.6 &$\pm$ 8.2 & $\pm$ 13.4 & $\pm$ 0.0 &  $\pm$ 16.0 & $\pm$ 16.8 & $\pm$ 10.8 & $\pm$ 6.0 \\
\midrule
\method{} &99.8 & 84.9& 95.2 & 86.6  & 100 & 78.3 & 80.9  & 99.2  & 91.8 \\
($N=4$) &$\pm$ 0.1 & $\pm$ 2.5 & $\pm$ 3.4 & $\pm$ 10.7 &$\pm$ 0.0 & $\pm$ 10.5 & $\pm$ 5.8 & $\pm$ 0.2 & $\pm$ 2.9 \\
\midrule
\method{} &99.8 & 95.2  & 98.9  & 92.6  & 100 & 91.3  & 89.4  & 99.2 & 95.8  \\
($N=8$)  & $\pm$ 0.1 & $\pm$ 2.4 & $\pm$ 0.9 & $\pm$ 3.7 & $\pm$ 0.0 & $\pm$ 5.9 & $\pm$ 5.6 & $\pm$ 0.6 & $\pm$ 1.2 \\
\midrule
\method{} &99.9 & 96.5 & 99.6 & 97.3 & 100  & 96.4 & 92.1  & 99.6  & 97.6  \\
($N=16$) &$\pm$ 0.0 & $\pm$ 1.0 & $\pm$ 0.4 & $\pm$ 1.0 & $\pm$ 0.0 & $\pm$ 0.7 & $\pm$ 3.0 & $\pm$ 0.1 & $\pm$ 0.5 \\
\midrule
\method{} &99.9 & 97.3 & 99.7 & 98.0  & 100  & 97.3  & 94.8  & 99.8 & 98.3 \\
($N=32$) &$\pm$ 0.0 & $\pm$ 0.7 & $\pm$ 0.1 & $\pm$ 0.5 & $\pm$ 0.0 & $\pm$ 0.5 &  $\pm$ 0.7 &  $\pm$ 0.0 &  $\pm$ 0.1 \\
\bottomrule
\end{tabular}
}
\end{table}

\begin{table}[t]
\centering
\caption{\textbf{Per-input inference latency and accuracy.} 
All results are obtained on computer vision tasks using CLIP ViT-B/32. 
\textbf{Res. proj.} denotes residual-projection latency, and \textbf{Inf.} denotes end-to-end inference latency.}
\label{tab:supp_efficiency_routing}
\footnotesize
\setlength{\tabcolsep}{3pt}

\begin{minipage}[t]{0.49\linewidth}
\centering
(a) Routing variants for TM-TIES+SiM
\resizebox{\linewidth}{!}{
\begin{tabular}{l|c|cc|cc}
\toprule
\textbf{\# Tasks} & \textbf{Res. proj.} & \textbf{Hard Inf.} & \textbf{Soft Inf.} & \textbf{Hard Acc.} & \textbf{Soft Acc.} \\
\midrule
8  & 0.006 ms & 0.003 s & 0.026 s & 92.7 & 89.2 \\
30 & 0.022 ms & 0.005 s & 0.039 s & 85.6 & 80.1 \\
\bottomrule
\end{tabular}
}
\end{minipage}
\hfill
\begin{minipage}[t]{0.49\linewidth}
\centering
(b) Dynamic model merging baselines
\resizebox{\linewidth}{!}{
\begin{tabular}{l|cc|cc}
\toprule
\textbf{Method} & \textbf{8-task Inf.} & \textbf{8-task Acc.} & \textbf{30-task Inf.} & \textbf{30-task Acc.} \\
\midrule
WEMoE~\cite{tang2024merging}       & 0.023 s & 90.4 & 0.035 s & 67.1 \\
MoW-Merging~\cite{ye2025dynamic} & 0.008 s & 88.1 & 0.013 s & 56.4 \\
\midrule
\method{} (hard)  & \textbf{0.003 s} & \textbf{92.7} & \textbf{0.005 s} & \textbf{85.6} \\
\bottomrule
\end{tabular}
}
\end{minipage}
\label{tab:eff}
\end{table}

\subsection{Varying rank $k$}
\label{app:ab_rank}
\cref{tab:k_ablation}  presents an ablation study investigating the impact of the rank $k$ on the task classification performance of \method{} across varying numbers of tasks (8, 14, and 20). 
Although our default configuration sets $k$ to 10\% of the feature dimension, the effective rank is inherently upper-bounded by the number of support samples per task ($N$).  Thus, in practice, the utilized rank is determined as $min(k,N)$.
According to these results,  \method{} demonstrates strong robustness to the choice of $k$, maintaining consistently high performance even with a very low rank.
Specifically, while increasing $k$ generally leads to a steady improvement in accuracy, the performance gains tend to saturate around $k=16$. Notably, even with the minimal rank of $k=1$, \method{} achieves competitive results across all settings, highlighting its efficiency and the effectiveness of the underlying representation. These findings underscore the practical reliability of \method{}, confirming that it is not overly sensitive to the specific value of $k$ and can achieve optimal performance with a low-rank approximation.

\subsection{Varying seeds and sample sizes} 
\label{app:ab_seed}
\cref{tab:seeds} presents an ablation study investigating the sensitivity of \method{} to two key factors: the sample size $N\in\{1,2,4,8,16,32\}$ and the random seed used for initialization.
While performance is understandably sensitive to seed selection in the extreme case of $N=1$, \method{} demonstrates rapid stabilization and enhanced robustness as $N$ increases. Specifically, with as few as 8 samples ($N=8$), the model already achieves high and robust performance across all tested seeds. As $N$ grows further, the accuracy continues to rise while the variance across different random seeds diminishes significantly, leading to nearly perfect and highly stable results for $N=32$. These findings underscore the practical reliability of \method{}, confirming its ability to provide consistent and highly robust performance even in data-limited few-shot scenarios where minimizing sensitivity to random variations is essential.

\begin{figure}[!t]
\centering
\begin{minipage}[t]{0.48\linewidth}
\centering
\includegraphics[width=\linewidth]{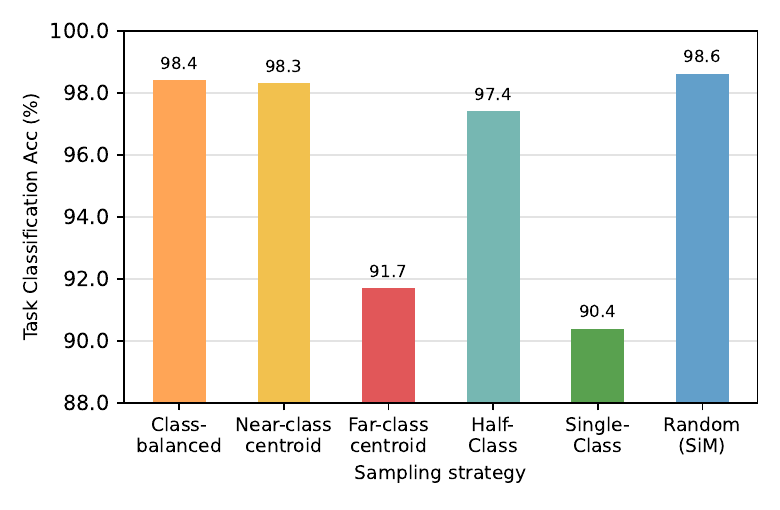}
\captionof{figure}{\textbf{Support-set quality analysis.}
Results are reported on 8 computer vision tasks using CLIP ViT-B/32.}
\label{fig:support_sampling_strategy}
\end{minipage}%
\hfill
\begin{minipage}[t]{0.48\linewidth}
\centering
\includegraphics[width=\linewidth]{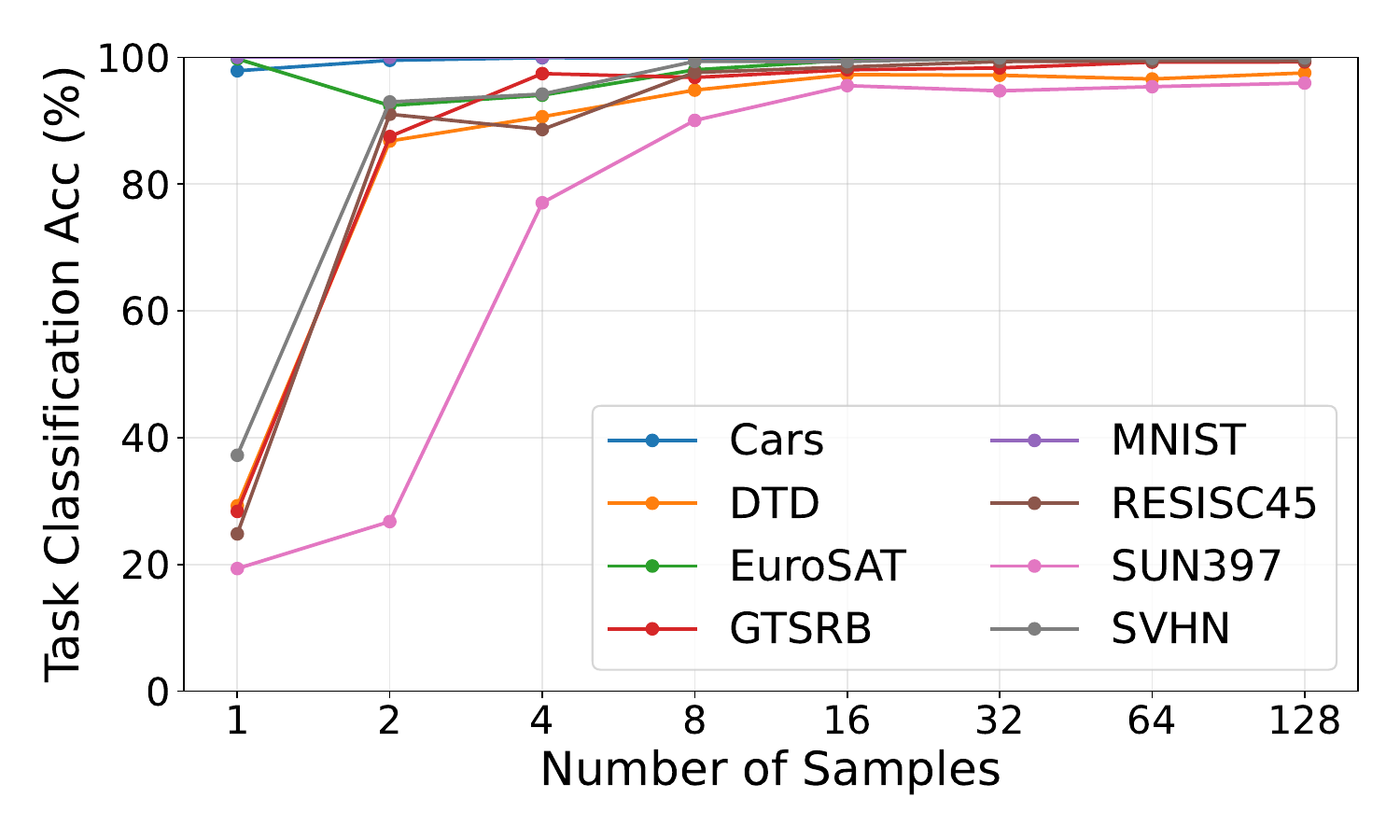}
\captionof{figure}{\textbf{Support-set quantity analysis.}
Task classification accuracy across different numbers of support samples.
Results are reported on 8 computer vision tasks using CLIP ViT-B/32.}
\label{fig:support_size_stability}
\end{minipage}
\end{figure}

\subsection{Support-set quantity and quality} 
\label{app:support-set}
We analyze the support set used for task subspace construction from two perspectives: sampling quality and support-set quantity.
We first evaluate the effect of support-set quality by varying the sampling strategy. 
As shown in~\cref{fig:support_sampling_strategy}, random sampling achieves the best performance, while class-balanced and near-class-centroid sampling obtain comparable accuracy. 
Even under biased support sets, such as far-class-centroid or single-class sampling, \method{} maintains accuracy above 90\%. 
This indicates that \method{} does not heavily depend on carefully curated support samples and remains robust to imperfect support-set selection.
We also examine how task classification accuracy changes with the number of support samples. 
As shown in~\cref{fig:support_size_stability}, \method{} rapidly improves as the support size increases and stabilizes with only a small number of samples across most datasets. Although SUN397~\cite{xiao2016sun} is more challenging in the extremely low-sample regime due to its large and diverse class space, the overall trend shows that \method{} can construct reliable task subspaces from few support samples.

\subsection{Inference efficiency and routing variants}
\label{app:efficiency}
We further analyze the inference efficiency of \method{} and justify our choice of hard routing. 
Although this residual-projection step scales linearly with the number of tasks, the absolute cost is extremely small.
As shown in~\cref{tab:eff}, the residual-projection latency is only 0.006 ms per input for 8 tasks and 0.022 ms per input for 30 tasks. 
Consequently, the end-to-end inference latency remains low even when the number of tasks increases, demonstrating that the routing overhead of \method{} is negligible in practice.
We also compare hard routing with a soft-routing alternative. 
Although soft routing can aggregate multiple task-specific paths for each input, this sample-wise aggregation prevents efficient group batching and substantially increases inference latency. 
In contrast, hard routing assigns each input to a single predicted task, enabling all inputs routed to the same task to be processed together. 
Accordingly, hard routing is both much faster and more accurate than soft routing in the 8-task and 30-task settings.
Moreover, hard routing does not discard multi-task knowledge, since \method{} is applied on top of task-specific masks or compressed task vectors produced by multi-task merging methods such as EMR and TALL-Mask. Thus, \method{} combines efficient group-batch inference with multi-task merged representations, providing a practical balance between accuracy and latency.

\begin{table}[t]
\centering
\caption{\textbf{Error propagation analysis.} 
Results are reported on 8 computer vision tasks using CLIP ViT-B/32. 
Oracle routing uses the ground-truth task identity, while \method{} routing uses the predicted task identity.}
\label{tab:error_propagation}
\small
\setlength{\tabcolsep}{5pt}
\renewcommand{\arraystretch}{1.12}

\begin{minipage}[t]{0.47\linewidth}
\centering
(a) Multi-task performance
\resizebox{\linewidth}{!}{
\begin{tabular}{l|cc}
\toprule
Method & Oracle routing & \method{} routing \\
\midrule
EMR~\cite{huang2024emr} & 91.3 & 90.7 \\
TM-TA~\cite{wang2024localizing} & 92.6 & 92.3 \\
TM-TIES~\cite{wang2024localizing} & 93.0 & 92.7 \\
TSV-C~\cite{gargiulo2024task} & 92.5 & 92.1 \\
\bottomrule
\end{tabular}}
\end{minipage}%
\hfill
\begin{minipage}[t]{0.47\linewidth}
\centering
(b) Error propagation for TM-TIES+\method{}
\resizebox{\linewidth}{!}{
\begin{tabular}{l|cc}
\toprule
 & Final correct & Final wrong \\
\midrule
Routing correct & 80,080 (93.4\%) & 5,616 (6.6\%) \\
Routing wrong & 980 (56.4\%) & 757 (43.6\%) \\
\bottomrule
\end{tabular}}
\end{minipage}
\end{table}

\subsection{Effect of routing errors on downstream performance} 
\label{app:routingerror}
We analyze how routing errors affect downstream multi-task performance.
As shown in~\cref{tab:error_propagation}, replacing \method{} routing with oracle routing yields only a small improvement of 0.3--0.6\% across different merging frameworks. 
This indicates that the routing errors of \method{} have limited impact on final accuracy.
We further inspect TM-TIES+\method{} and find that 56.4\% of incorrectly routed samples still produce correct final predictions. This suggests that routing errors do not necessarily propagate to prediction errors, partly because misrouting often occurs between semantically related tasks. 
For example, among 32 SVHN~\cite{netzer2011reading} samples misrouted to MNIST~\cite{deng2012mnist}, 27 samples are still correctly classified.
Thus, \method{} remains robust at the downstream prediction level even when task routing is imperfect.

\section{Limitations}
\label{app:limitations}

Despite its effectiveness, \method{} has certain limitations. 
Primarily, it requires a small set of samples to construct task-specific subspaces.
However, our method offers a more efficient alternative to other dynamic merging approaches~\cite{lu2024twin,tang2024merging,ye2025dynamic} that often require extensive training phases (with a substantial amount of data) to optimize routers or gating networks.
Furthermore, while other dynamic merging approaches require access to data during model merging process, our method requires a small set of data prior to model merging process.
In fact, task-specific subspaces can be prepared together with task experts.
Thus, our method does not need to maintain data storage even after task expert release or during model merging process.
On the other hand, other router-based dynamic model merging methods need to store data until model merging process.
Regardless, the data usage can be a limitation, and removing the need for data and constructing task manifolds is an interesting yet practical research direction to pursue for future research.


\end{document}